\documentclass[default, iicol, sn-mathphys-num]{sn-jnl}

\usepackage{graphicx}%
\usepackage{multirow, multicol}
\usepackage{amsmath,amssymb,amsfonts}%
\usepackage{amsthm}%
\usepackage{mathrsfs}%
\usepackage[title]{appendix}%
\usepackage{textcomp}%
\usepackage{manyfoot}%
\usepackage{booktabs}%
\usepackage{algorithm}%
\usepackage{algorithmicx}%
\usepackage{algpseudocode}%
\usepackage{listings}%
\usepackage{tabularx}

\usepackage{mathtools}
\usepackage{adjustbox}
\usepackage{array}
\usepackage{makecell}
\usepackage{boldline}
\usepackage{pifont}%xmark
\usepackage{subfig}
\usepackage{subcaption}
\usepackage{diagbox}
\usepackage{graphicx}

\usepackage{algorithm}
\usepackage{algpseudocode}
\usepackage{caption}
\usepackage{amsmath}
\usepackage{amssymb}
\usepackage{siunitx}
\usepackage[table]{xcolor} 
\definecolor{lightblue}{RGB}{220,230,241} 
\usepackage[marginal]{footmisc}

\theoremstyle{thmstyleone}%
\theoremstyle{thmstyletwo}%

\theoremstyle{thmstylethree}%

\begin{document}

\title[Article Title]{Task-Specific Knowledge Distillation from the Vision Foundation Model for Enhanced Medical Image Segmentation}


\author[1,2]{\fnm{Pengchen} \sur{Liang}\texorpdfstring{$^{\dagger}$}{}}

\author[3]{\fnm{Haishan} \sur{Huang}\texorpdfstring{$^{\dagger}$}{}}

\author[4]{\fnm{Bin} \sur{Pu}\texorpdfstring{$^{\dagger}$}{}}

\author*[3]{\fnm{Jianguo} \sur{Chen}}\email{chenjg33@mail.sysu.edu.cn}

\author[3]{\fnm{Xiang} \sur{Hua}}

\author[5]{\fnm{Jing} \sur{Zhang}}

\author[6]{\fnm{Weibo} \sur{Ma}}

\author[4]{\fnm{Zhuangzhuang} \sur{Chen}}

\author*[7]{\fnm{Yiwei} \sur{Li}}\email{liyiwei@shchildren.com.cn}

\author*[1]{\fnm{Qing} \sur{Chang}}\email{robie0510@hotmail.com}

\affil[1]{\orgdiv{Department Shanghai Key Laboratory of Gastric Neoplasms, Department of Surgery, Shanghai Institute of Digestive Surgery, Ruijin Hospital}, \orgname{Shanghai Jiao Tong University School of Medicine}, \orgaddress{\city{Shanghai}, \postcode{200025}, \country{China}}}

\affil[2]{\orgdiv{School of Microelectronics}, \orgname{Shanghai University}, \orgaddress{\city{Shanghai}, \postcode{201800}, \country{China}}}

\affil[3]{\orgdiv{School of Software Engineering}, \orgname{Sun Yat-sen University}, \orgaddress{\city{Zhuhai}, \postcode{519000}, \country{China}}}

\affil[4]{\orgdiv{Department of Electronic and Computer Engineering}, \orgname{The Hong Kong University of Science and Technology}, \orgaddress{\city{Hong Kong}, \state{SAR}, \country{China}}}

\affil[5]{\orgdiv{Department of Radiology, Ruijin Hospital}, \orgname{Shanghai Jiaotong University School of Medicine}, \orgaddress{\city{Shanghai}, \postcode{200025}, \country{China}}}

\affil[6]{\orgdiv{School of Public Administration}, \orgname{East China Normal University}, \orgaddress{\city{Shanghai}, \postcode{200062}, \country{China}}}

\affil[7]{\orgdiv{Department of Nuclear Medicine, Shanghai Children's Hospital, School of Medicine}, \orgname{Shanghai Jiao Tong University}, \orgaddress{\city{Shanghai}, \postcode{200062}, \country{China}}}
% \footnotetext[1]{$^{\dagger}$(Co-first authors)}



\abstract{
Large-scale pre-trained models, such as Vision Foundation Models (VFMs), have demonstrated impressive performance across various downstream tasks by transferring generalized knowledge, especially when target data is limited. 
However, their high computational cost and the domain gap between natural and medical images limit their practical application in medical segmentation tasks. Motivated by this, we pose the following important question: "How can we effectively utilize the knowledge of large pre-trained VFMs to train a small, task-specific model for medical image segmentation when training data is limited?" To address this problem, we propose a novel and generalizable task-specific knowledge distillation framework.
Our method fine-tunes the VFM on the target segmentation task to capture task-specific features before distilling the knowledge to smaller models, leveraging Low-Rank Adaptation (LoRA) to reduce the computational cost of fine-tuning. 
Additionally, we incorporate synthetic data generated by diffusion models to augment the transfer set, enhancing model performance in data-limited scenarios.
Experimental results across five medical image datasets demonstrate that our method consistently outperforms task-agnostic knowledge distillation and self-supervised pretraining approaches like MoCo v3 and Masked Autoencoders (MAE). 
For example, on the KidneyUS dataset, our method achieved a 28\% higher Dice score than task-agnostic KD using 80 labeled samples for fine-tuning. On the CHAOS dataset, it achieved an 11\% improvement over MAE with 100 labeled samples.
These results underscore the potential of task-specific knowledge distillation to train accurate, efficient models for medical image segmentation in data-constrained settings.
}


\maketitle
\footnote{\texorpdfstring{$^{\dagger}$}{}Pengchen Liang, Haishan Huang, and Bin Pu contributed equally to this work.}

\section{Introduction}\label{sec1}

Medical image segmentation is essential for clinical diagnosis, treatment planning, and surgical guidance, as it enables precise delineation of anatomical structures and pathological regions~\cite{liang2023dawtran, abut2023paradigm, munia2024attention, li2024segmentation, pu2021automatic}. Despite advances in deep learning, developing high-performance segmentation models remains challenging due to the scarcity of annotated medical data. Annotating medical images is costly and time-intensive, requiring specialized expertise to label complex structures~\cite{tajbakhsh2020embracing} meticulously. This scarcity hampers the training of effective models, which typically require large amounts of labeled data to generalize well.

Vision Foundation Models (VFMs), pre-trained on extensive and diverse datasets, have emerged as promising tools by capturing generalized features transferable to various downstream tasks~\cite{oquab2023dinov2, kirillov2023segment}. These models have shown potential in medical image segmentation through transfer learning~\cite{chen2024ma, zhang2023input, wang2024sam, leng2024self}. However, directly applying VFMs to medical imaging faces two significant challenges. \textbf{\emph{(1) The domain gap between natural images, on which VFMs are typically trained, and medical images limit their effectiveness due to differences in features and structures~\cite{raghu2019transfusion, atasever2023comprehensive, messaoudi2023cross}.} \emph{(2) VFMs are often large and computationally intensive, making them impractical for deployment in resource-constrained clinical environments.}}

\begin{figure*}[ht]
    \centering
    \includegraphics[width=\textwidth]{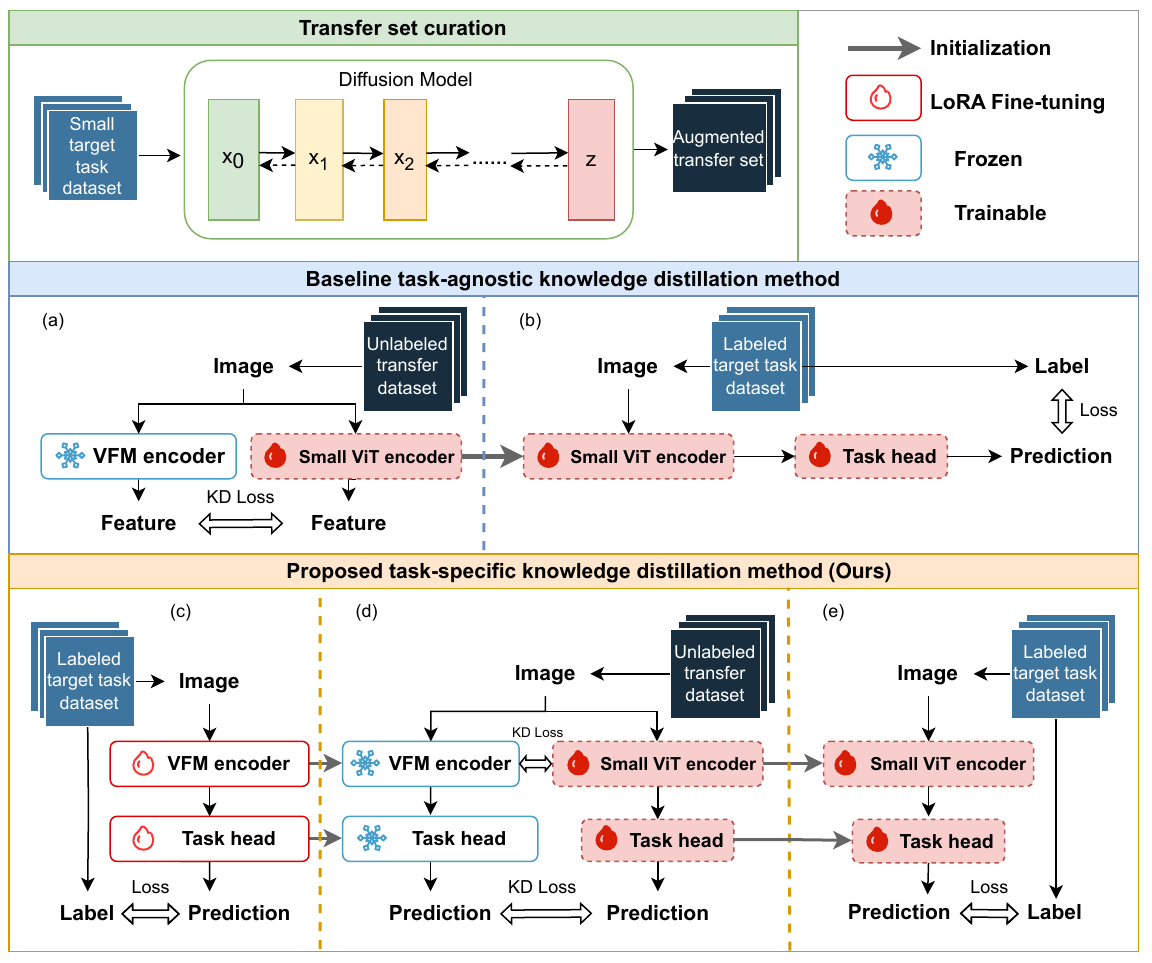}
    \caption{Overview of the proposed task-specific knowledge distillation framework compared with the task-agnostic method. \textbf{Top:} Transfer set curation using a diffusion model to generate synthetic data from a small labeled dataset, expanding the training set. \textbf{Middle:} \textit{Task-agnostic knowledge distillation method:} (a) The small Vision Transformer (ViT) model is pre-trained by matching its features to those extracted by the VFM on the unlabeled transfer set, focusing on general feature alignment. (b) The small ViT is fine-tuned using the labeled target task data. \textbf{Bottom:} \textit{Proposed task-specific knowledge distillation method:} (c) The VFM is first fine-tuned on the target task using LoRA adaptation. (d) The small ViT model is pre-trained by aligning hidden layer representations and segmentation-specific predictions between the VFM and the small ViT on the unlabeled transfer set. (e) The small ViT model is fine-tuned using the labeled target task data to optimize segmentation performance.}
    \label{fig:knowledge_transfer}
\end{figure*}

An effective solution is to transfer the knowledge of VFMs to smaller, task-specific models that are efficient and tailored for medical image segmentation. Knowledge distillation (KD) provides a framework for this transfer, where a large "teacher" model imparts its knowledge to a smaller "student" model~\cite{hinton2015distilling, gou2021knowledge, li2024lorkd, liang2024rskd, zhao2023mskd}. 
Previous typical task-agnostic KD methods (Figure~\ref{fig:knowledge_transfer}(Middle)) often focus on transferring generic feature representations without tailoring to the specific needs of the target task~\cite{xiong2024efficientsam}. 
These approaches may inadequately capture crucial features necessary for accurate segmentation, such as fine anatomical boundaries and subtle pathological variations, reducing the effectiveness of the smaller model.

To address these challenges, we propose a novel task-specific knowledge distillation framework (Figure~\ref{fig:knowledge_transfer}(Bottom)) that effectively transfers knowledge from large pre-trained VFMs to smaller, efficient models tailored for medical image segmentation. Our framework first fine-tunes the VFM on the target segmentation task using the limited available labeled medical data. This fine-tuning enables the VFM to learn task-relevant features specific to medical images. We then distill this task-specific knowledge to a smaller model by aligning both the intermediate representations and the final segmentation outputs between the fine-tuned VFM and the small model. This process ensures that the small model inherits the critical features necessary for high-accuracy segmentation, overcoming the limitations of traditional KD approaches.

To reduce the computational cost associated with fine-tuning large VFMs, we employ Low-Rank Adaptation (LoRA) ~\cite{hu2021lora, yan2024after, cheng2024unleashing, wang2024task}, which allows efficient adaptation of large models with minimal additional parameters. Furthermore, recognizing the scarcity of medical data, we leverage diffusion models~\cite{ho2020denoising, kawar2022denoising, garcea2023data} to generate synthetic medical images. These synthetic images augment the unlabeled transfer set (Figure~\ref{fig:knowledge_transfer}(Top)) used during the knowledge distillation process, enhancing the diversity of training samples and enabling the small model to learn more effectively from the fine-tuned VFM in data-limited scenarios.

In our experiments, we also compare the effectiveness of our task-specific KD framework with self-supervised learning (SSL) approaches, which have emerged as an alternative to mitigate the need for large amounts of labeled data. SSL-based methods, such as Momentum Contrast (MoCo v3)\cite{chen2021empirical} and Masked Autoencoders (MAE)\cite{he2022masked}, allow models to learn from unlabeled data by solving pretext tasks to learn useful representations. While these methods have demonstrated success in medical image segmentation~\cite{zheng2021hierarchical, miao2023sc, pani2024hybrid, yan2023label}, we show that our task-specific KD framework consistently outperforms SSL methods in data-limited scenarios.

Our key contributions are as follows:

\begin{itemize}

\item We propose a task-specific KD framework that leverages fine-tuned VFM to transfer domain-specific knowledge to smaller Vision Transformer (ViT) models. By aligning hidden layer representations and segmentation outputs, our method achieves more comprehensive and accurate knowledge transfer for medical image segmentation, overcoming the limitations of traditional KD approaches.

\item To reduce the computational cost of fine-tuning large VFMs, we adopt LoRA, allowing for efficient adaptation of the VFM while preserving its large-scale pre-training benefits.

\item We enhance the KD process by generating an augmented transfer set with diffusion models, increasing the diversity and size of the training data. Our analysis of different transfer set sizes (1,000; 2,000; and 3,000 synthetic images) shows that task-specific KD scales effectively with larger transfer sets, leading to significant performance improvements in data-limited segmentation tasks.

\item We conduct a thorough evaluation of task-specific KD against state-of-the-art techniques, including task-agnostic KD and self-supervised pretraining methods such as MoCo v3 and MAE. Our experiments span various data-constrained settings, providing new insights into the relative performance of each approach across different medical segmentation tasks.

\item Through extensive experiments on five diverse medical imaging datasets, we demonstrate that task-specific KD consistently outperforms competing methods, particularly in data-limited conditions. Notably, our approach achieves a 74.82\% Dice score improvement in retinal vessel segmentation on the DRIVE dataset and a 13.59\% improvement in kidney segmentation on the KidneyUS dataset, highlighting its robustness in complex segmentation tasks.

\end{itemize}

\vspace{-4pt}
\section{Related Work}
\label{sec:related_work}
%
% \vspace{-4pt}

\subsection{Knowledge Distillation (KD)}
Knowledge distillation (KD) has emerged as an effective approach for transferring knowledge from a large teacher model to a smaller student model, thereby reducing computational requirements while retaining performance. Various KD strategies have been explored over the years, including transferring logits from teacher to student~\cite{hinton2015distilling}, matching intermediate feature representations~\cite{zhang2020task, heo2019comprehensive}, and utilizing attention maps~\cite{liu2019attention,guo2023class}. In the medical imaging domain, these methods have shown promise for enhancing performance, especially when dealing with complex tasks such as segmentation. However, the use of task-specific KD for transferring detailed anatomical information remains under-explored, particularly in scenarios involving scarce labeled data. 
Recent research has also explored distillation in multimodal settings such as image-language models~\cite{zhang2024vision, sameni2024building}, but fewer works focus on the unique challenges presented by medical image segmentation. Our work investigates task-specific KD approaches, extending these traditional KD strategies by leveraging VFM, which is fine-tuned with task-relevant data for improved medical image segmentation.

\subsection{Self-Supervised Learning and Transfer Learning}
Self-supervised learning (SSL) has gained significant attention in medical image analysis, particularly due to the scarcity of labeled data. SSL methods, such as contrastive learning (e.g., MoCo v3~\cite{chen2021empirical}) and reconstruction-based approaches like Masked Autoencoders (MAE)~\cite{he2022masked}, have enabled models to learn useful representations from abundant unlabeled medical imaging data. 
SSL helps capture the underlying structures present in medical images, which is crucial for downstream tasks like segmentation. Transfer learning has also been extensively used in medical imaging, typically involving pretraining models on large-scale natural image datasets such as ImageNet~\cite{kornblith2019better}. 
Nevertheless, the domain gap between natural images and medical images can limit the generalization capabilities of these models~\cite{su2023rethinking}. 
By combining SSL with KD, our work leverages both representation learning from unlabeled datasets and task-specific knowledge transfer to achieve robust performance in medical image segmentation.

\subsection{Task-Specific Knowledge Transfer in Vision Foundation Model}
Vision Foundation Model (VFM), trained on diverse large-scale datasets, represents a significant advancement in image understanding across various domains, including medical imaging. Recent efforts, such as DINOv2~\cite{oquab2023dinov2}, have explored transferring knowledge from VFM in a task-agnostic manner, often using methods designed for transformer architectures. 
Task-specific knowledge transfer, however, has received less attention in medical segmentation tasks, where domain-specific knowledge is vital for model accuracy. In our study, we focus on transferring knowledge from VFM that has been fine-tuned on specific segmentation tasks using methods like LoRA~\cite{hu2021lora,zhang2023customized}. 
We present a task-specific KD approach where the VFM is first fine-tuned on medical data, and the distilled knowledge is then transferred to a smaller Vision Transformer (ViT). Unlike traditional KD which often focuses on feature extraction, our approach aligns both feature representations and target predictions, ensuring a more comprehensive knowledge transfer. By constructing and evaluating transfer sets of different sizes, we demonstrate the benefits of this approach for efficient training in medical image segmentation, especially in data-scarce settings.

\section{Methodology}
\subsection{Overview}
\label{sec:method}
In this work, we propose a generalizable task-specific knowledge distillation (TS-KD) framework for medical image segmentation, addressing data scarcity, efficient model adaptation, and comprehensive knowledge transfer. (1) A Swin-transformer-based diffusion model generates synthetic datasets to augment limited medical imaging data. (2) we fine-tune the Segment Anything Model (SAM) using Low-Rank Adaptation (LoRA), which efficiently adapts SAM’s parameters for task-specific segmentation needs. (3) Our dual-level distillation process aligns feature representations and segmentation outputs between SAM and a ViT-Tiny model, ensuring the effective transfer of both general and task-specific knowledge for improved segmentation performance in compact models.

\subsection{Transfer Dataset Generation}

To address the challenge of data scarcity in medical imaging, we used a Swin-transformer-based diffusion model to generate synthetic transfer datasets (Figure \ref{fig:knowledge_transfer}(TOP)). Each of the five primary datasets was augmented using data augmentation techniques (rotation, scaling, and affine transformations) to create a dataset of 1,000 images per task. These augmented datasets were then used to train the diffusion model for generating additional synthetic images.

The diffusion model is based on the research~\cite{pan20232d}, utilizing a Swin-transformer-based network that employs a forward Gaussian noise process and a reverse denoising process. The forward process gradually adds noise to the input images over multiple time steps:

\begin{equation}
q(x_t | x_{t-1}) = \mathcal{N}(x_t; \sqrt{\alpha_t} x_{t-1}, (1 - \alpha_t)I),
\end{equation}
where $x_t$ represents the noisy image at time step $t$, and $\alpha_t$ is the noise scaling factor. The reverse process attempts to predict and denoise the noisy images to generate realistic synthetic images:

\begin{equation}
p_{\theta}(x_{t-1} | x_t) = \mathcal{N}(x_{t-1}; \mu_{\theta}(x_t, t), \sigma_t^2 I),
\end{equation}
where $\mu_{\theta}(x_t, t)$ is the predicted mean by the model with parameters $\theta$, and $\sigma_t^2$ is the variance of the noise.

\subsection{Task-Specific Knowledge Distillation (KD)}

\begin{figure*}[htbp]
    \centering
    \includegraphics[scale=0.71]{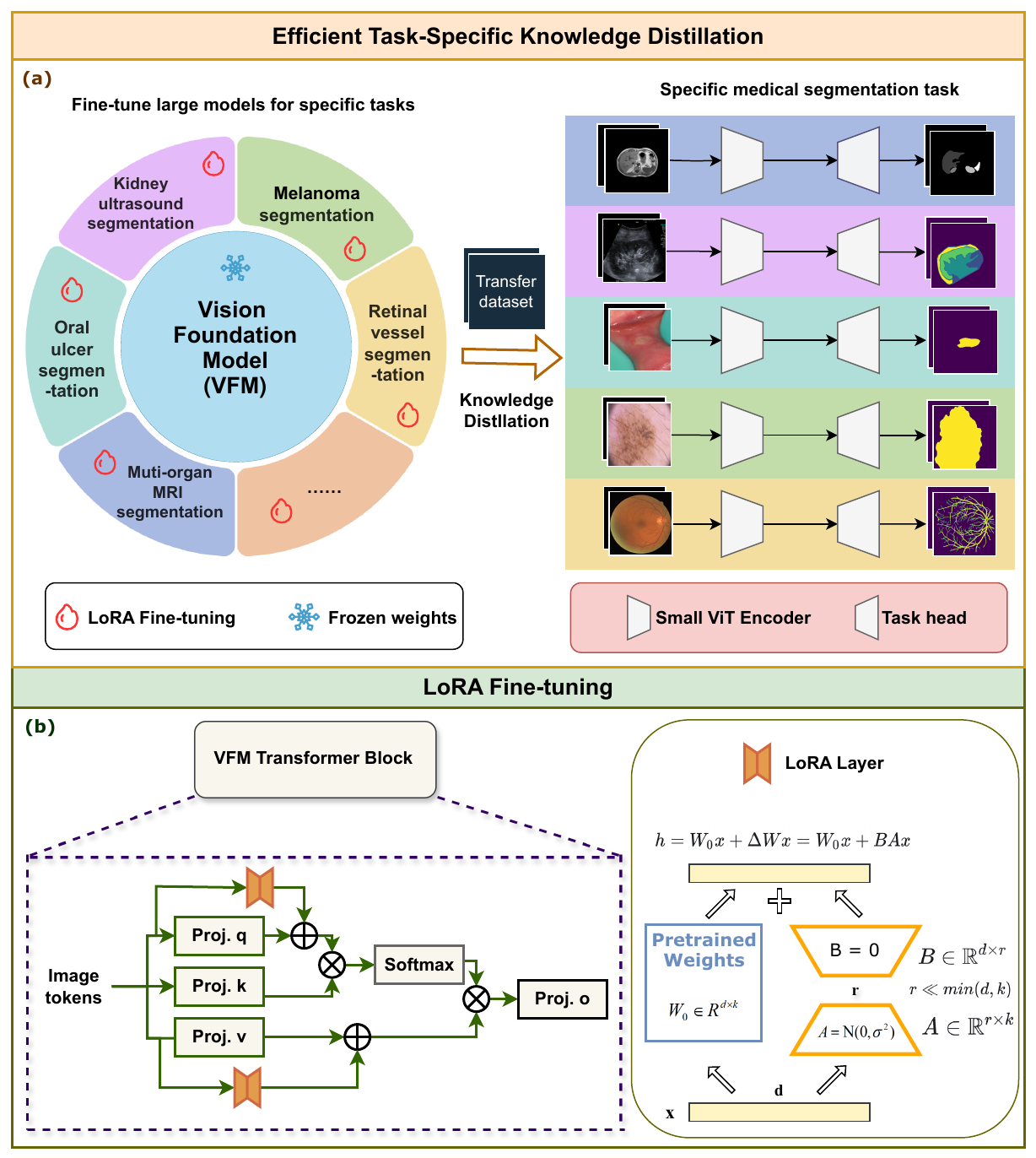}
    \caption{Efficient Task-Specific Knowledge Distillation framework. (a) Fine-tuning large models for specific medical segmentation tasks, such as kidney ultrasound, melanoma, and retinal vessel segmentation, using the Vision Foundation Model ('Segment Anything Model'). The framework utilizes LoRA fine-tuning to adapt VFM's encoder and decoder for these specific tasks. Transfer datasets are used in combination with knowledge distillation to train smaller task-specific models for efficient deployment. (b) Detailed architecture of LoRA fine-tuning, showing the VFM transformer's block with LoRA layers for efficient parameter updates, where pre-trained weights are kept frozen while low-rank matrices adjust the attention mechanism to fit the target task.}
    \label{fig:KD-LoRA}
\end{figure*}

As shown in Figure \ref{fig:KD-LoRA}, we proposed a Task-Specific Knowledge Distillation (KD) framework combined with LoRA Fine-Tuning to effectively transfer knowledge from a large SAM model to a smaller ViT-Tiny model for segmentation tasks. This approach aims to distill both general and task-specific knowledge, ensuring that the student model learns both the encoder’s feature extraction capabilities and the decoder’s task-specific outputs.

\subsubsection{LoRA Fine-Tuning of the SAM Model}

To adapt the large SAM model to the target segmentation task, we use LoRA (Low-Rank Adaptation) for fine-tuning both the SAM encoder and decoder(as illustrated in Figure \ref{fig:KD-LoRA}b). LoRA enables efficient adaptation by injecting trainable low-rank matrices into the model’s attention layers, reducing the number of parameters that need to be updated while preserving most of the original model's knowledge. 

For each attention weight matrix $W_0 \in \mathbb{R}^{d \times d}$ in the SAM model, LoRA introduces two low-rank matrices $A \in \mathbb{R}^{d \times r}$ and $B \in \mathbb{R}^{r \times d}$, where $r \ll d$, and modifies the original weight matrix as follows:

\begin{equation}
W = W_0 + \Delta W, \quad \Delta W = A \cdot B,
\end{equation}
where $\Delta W$ is the task-specific adaptation, and $r$ is a hyperparameter controlling the adaptation's rank. This technique allows us to fine-tune only the added low-rank matrices $A$ and $B$, while keeping the original weights $W_0$ frozen, enabling efficient task-specific adaptation of the SAM model.

\paragraph{Detail teachnique of LoRA.}
The LoRA fine-tuning was conducted on both the query and value projection layers of each transformer block within the SAM model's encoder. Freezing the remaining parameters of the SAM model ensures minimal update overhead. Following settings in \cite{zhang2023customized}, the AdamW optimizer was used with an initial learning rate of 0.005, and the fine-tuning process utilized a warmup strategy to stabilize the early training stages. The warmup period was set to 250 iterations, with a maximum of 160 epochs for training. The cross-entropy loss and Dice loss were combined in a $0.2:0.8$ ratio to supervise the fine-tuning process:

\begin{equation}
L = \lambda_1 \text{CE}(S, \hat{S}) + \lambda_2 \text{Dice}(S, \hat{S}),
\end{equation}
where $\lambda_1 = 0.2$ and $\lambda_2 = 0.8$ control the balance between the two loss components.

\subsubsection{Dual-Level Knowledge Distillation}

To transfer knowledge from the fine-tuned SAM model to the ViT-Tiny model, we adopt a dual-level knowledge distillation strategy that aligns both the encoder’s hidden representations and the decoder’s output predictions.

\paragraph{Encoder in KD.}
We match the hidden representations from the final layers of the encoders in the student and teacher models, allowing the student model (ViT-Tiny) to learn from the rich feature representations of the SAM model. The distillation loss is given by:

\begin{equation}
\mathcal{L}_{KD}^{encoder} = \frac{1}{N} \sum_{i=1}^N \|h_i^{(student)} - h_i^{(teacher)}\|^2,
\end{equation}
where $N$ is the number of feature map elements, and $h_i^{(student)}$ and $h_i^{(teacher)}$ are the hidden states from the student and teacher encoders, respectively.

\paragraph{Decoder in KD.}
To transfer task-specific knowledge, we also align the outputs of the student and teacher decoders. Specifically, the output logits of the SAM decoder (producing low-resolution segmentation masks) are matched with the final output of the ViT-Tiny decoder. The distillation loss for this process is given by:

\begin{equation}
\mathcal{L}_{KD}^{decoder} = \frac{1}{M} \sum_{j=1}^M \|y_j^{(student)} - y_j^{(teacher)}\|^2,
\end{equation}
where $y_j^{(student)}$ and $y_j^{(teacher)}$ represent the predicted segmentation logits, and $M$ is the number of elements (pixels) in the segmentation mask.

\subsection{Theoretical Derivations}
In this section, we present a mathematical analysis to substantiate the theoretical advantages of Task-Specific Knowledge Distillation (TS-KD) over Task-Agnostic Knowledge Distillation (TA-KD). By employing principles from PAC learning theory and information theory, we demonstrate how TS-KD leads to reduced generalization error and greater sample efficiency, especially in data-scarce scenarios like medical image segmentation.

\subsubsection{Generalization Error in Knowledge Distillation}
Given a student model $F_S$ trained via distillation from a teacher model $F_T$, the generalization error can be expressed as:
\begin{equation}
R(F_S) = \mathbb{E}_{(x, y) \sim \mathcal{D}}[\ell(F_S(x), y)],
\end{equation}
where $\ell(\cdot, \cdot)$ is the task-specific loss function (e.g., cross-entropy or Dice loss). In knowledge distillation, the goal is to minimize the divergence between the student and teacher outputs while also ensuring that the student model generalizes well to the underlying task. The total error can be decomposed as:
\begin{equation}
R(F_S) = \text{Bias}^2(F_S) + \text{Variance}(F_S).
\end{equation}

\subsubsection{Hypothesis Space and VC Dimension}
In distillation, the complexity of the hypothesis space $\mathcal{H}$ that the student model explores plays a crucial role in determining generalization performance. Let the VC dimension $\text{VCdim}(\mathcal{H})$ represent the complexity of the hypothesis space. For a given training set size $m$, PAC learning bounds the generalization error as:
\begin{equation}
R(F_S) \leq \hat{R}_m(F_S) + O\left(\sqrt{\frac{\text{VCdim}(\mathcal{H})}{m}}\right),
\end{equation}
where $\hat{R}_m(F_S)$ is the empirical error on the training data. In TS-KD, since the teacher model $F_T^{\text{LoRA}}$ has been fine-tuned for the specific task, the student model's hypothesis space $\mathcal{H}_{\text{TS}}$ is naturally more restricted, leading to a lower VC dimension compared to TA-KD:
\begin{equation}
\text{VCdim}(\mathcal{H}_{\text{TS}}) < \text{VCdim}(\mathcal{H}_{\text{TA}}).
\end{equation}
This reduction in hypothesis space complexity directly implies better generalization for TS-KD, as the generalization bound becomes tighter with fewer training examples.

\subsubsection{Sample Complexity}
In PAC learning, the sample complexity required to achieve a generalization error of at most $\epsilon$ with high probability $1 - \delta$ is given by:
\begin{equation}
m \geq O\left(\frac{1}{\epsilon^2} \left( \text{VCdim}(\mathcal{H}) + \ln\frac{1}{\delta} \right)\right).
\end{equation}
Since $\text{VCdim}(\mathcal{H}_{\text{TS}}) < \text{VCdim}(\mathcal{H}_{\text{TA}})$, TS-KD requires fewer training examples to reach the same level of generalization, making it particularly advantageous when dealing with limited labeled data, as is common in medical imaging tasks.

\subsubsection{Information-Theoretic Analysis}
From an information-theoretic perspective, the effectiveness of knowledge distillation can be evaluated through the lens of information gain. The mutual information between the teacher’s output and the true label $y$ provides insight into how well the teacher’s predictions guide the student:
\begin{equation}
I(T; Y) = H(Y) - H(Y | T),
\end{equation}
In TA-KD, the teacher $F_T$ is not fine-tuned for the task, which increases the entropy $H(T)$ of its output. Consequently, the mutual information $I(T; Y)$ is lower compared to TS-KD, where the teacher model $F_T^{\text{LoRA}}$ is task-specific:
\begin{equation}
I(T^{\text{LoRA}}; Y) > I(T; Y).
\end{equation}
This implies that TS-KD provides greater information gain for the student model, leading to better task performance and lower generalization error.

Through PAC learning and information-theoretic analysis, we have demonstrated that TS-KD offers significant theoretical advantages over TA-KD. By reducing the hypothesis space complexity and increasing the mutual information between the teacher's output and the true label, TS-KD ensures better generalization with fewer training examples, making it particularly suitable for data-limited domains such as medical imaging.
\section{Experiment}
\label{sec:experiment}

\subsection{Experimental Setup}

To comprehensively evaluate the effectiveness of the proposed task-specific knowledge distillation (KD) strategy for medical image segmentation, we designed experiments involving six distinct model groups, assessing various pre-training and fine-tuning strategies for lightweight models. This section describes the experimental setup, including datasets, model configurations, procedures, and evaluation metrics.

\subsubsection{Datasets}

We utilized five primary datasets to address different segmentation challenges in medical imaging, covering diverse modalities such as ultrasound, oral cavity, abdominal, dermoscopic, and retinal images. These datasets enable a robust evaluation of our method's generalizability and effectiveness:

\begin{itemize} 

\item \textbf{KidneyUS Dataset~\cite{singla2023open}}: This ultrasound dataset focuses on kidney segmentation, with pixel-wise annotations delineating kidney regions. The images, sized at $512 \times 512$, present challenges due to ultrasound noise and anatomical variability, testing the robustness of segmentation models in a low-contrast setting.

\item \textbf{Autooral Dataset~\cite{jiang2024high}}: This dataset provides images of the oral cavity for ulcer segmentation, with masks marking ulcerated regions. With $256 \times 256$ resolution, it challenges models with varying lighting conditions and complex anatomical structures within the oral cavity.

\item \textbf{CHAOS Dataset~\cite{kavur2021chaos}}: A benchmark dataset for multi-organ segmentation in abdominal MRI, containing organs like the liver, kidneys, and spleen, at $256 \times 256$ resolution. It assesses models’ adaptability in segmenting organs with complex shapes and low-contrast boundaries.

\item \textbf{PH2 Dataset~\cite{mendoncca2015ph2}}: This dermoscopic dataset, with images at $512 \times 512$, is designed for skin lesion segmentation, including melanomas. The dataset provides a high-resolution view, essential for capturing fine details in texture and asymmetry to distinguish pathological regions.

\item \textbf{DRIVE Dataset~\cite{staal2004ridge}}: A retinal fundus image dataset used for vessel segmentation, providing $512 \times 512$ images with vascular structure labels. This dataset evaluates the model's precision in segmenting intricate vascular patterns, critical for detecting eye conditions like diabetic retinopathy.

\end{itemize}

\subsubsection{Model Architectures}

The models used in our experiments were based on Vision Transformer (ViT) architectures:

\begin{itemize}
    \item \textbf{Vision Foundation Model (VFM).} We adopted the ViT-B model from SAM as the teacher model for KD. This model is a vision foundation model pre-trained on diverse datasets to capture robust visual representations. For task-specific knowledge distillation, we fine-tuned the SAM model on the labeled portion of each target dataset using Low-Rank Adaptation (LoRA).
    
    \item \textbf{Small ViT Model.} The student model was based on a ViT-Tiny~\cite{dosovitskiy2020vit} architecture, which was designed to be resource-efficient for deployment. The segmentation head in this model was built using a Feature Pyramid Network (FPN). The FPN takes the output from the last layer of the ViT-Tiny encoder, performs multi-scale feature aggregation, and generates the segmentation output. This model served as the student in KD and underwent multiple training strategies, including training from scratch, self-supervised pre-training, and knowledge distillation.
\end{itemize}

\subsubsection{Baseline Pre-Training Strategies}

We conducted experiments across different training groups to evaluate various pre-training techniques:

\begin{enumerate}
    \item \textbf{ImageNet Pre-trained Small ViT Model with MAE.} Instead of training the model from scratch, we used pre-trained weights from the work~\cite{wang2023closer}, where the ViT tiny model was pre-trained on the ImageNet dataset using Masked Autoencoders (MAE). These pre-trained weights were then fine-tuned on our labeled target dataset. Due to the difference in input image size, we did not load the first layer of tokens when initializing the ViT-Tiny model, as the number of tokens differed from those in the original pre-training. 
    
    \item \textbf{Self-Supervised Pre-training with MoCo v3.} We applied Momentum Contrast (MoCo v3) for self-supervised pre-training on an unlabeled transfer dataset of varying sizes ($1,000$, $2,000$, and $3,000$ images). The MoCo v3 objective involves maintaining consistency between a query and a momentum-averaged key encoder:
    \begin{equation}
    \mathcal{L}_{MoCo} = -\log \frac{\exp(q \cdot k_+ / \tau)}{\sum_{i=0}^K \exp(q \cdot k_i / \tau)},
    \end{equation}
    where $q$ is the query, $k_+$ is the positive key, $k_i$ are negative keys, and $\tau$ is the temperature parameter.

    \item \textbf{Self-Supervised Pre-training with MAE.} In this group, we employed Masked Autoencoders for self-supervised pre-training, following the same transfer datasets as in the MoCo experiments. This allowed us to compare the effectiveness of different self-supervised learning techniques.
    The MAE pre-training objective involves reconstructing the original image from a masked version:
    \begin{equation}
    \mathcal{L}_{MAE} = \frac{1}{N} \sum_{i=1}^N \|x_i - f_{\theta}(\tilde{x}_i)\|^2,
    \end{equation}
    where $x_i$ is the original image, $\tilde{x}_i$ is the masked image, and $f_{\theta}$ is the MAE model.

    \item \textbf{Task-Agnostic Knowledge Distillation (KD).} In this approach, we performed knowledge distillation from only the SAM encoder to the ViT-Tiny encoder. The distillation loss function minimizes the mean squared error (MSE) between the hidden representations from the last hidden layers of the SAM and ViT-Tiny encoders:
    \begin{equation}
    \mathcal{L}_{KD}^{encoder} = \frac{1}{N} \sum_{i=1}^N \|h_i^{(student)} - h_i^{(teacher)}\|^2,
    \end{equation}
    where $h_i^{(student)}$ and $h_i^{(teacher)}$ represent the hidden representations from the student and teacher encoders, respectively.

\end{enumerate}

\subsection{Training Procedure}

For all training strategies, we followed a consistent procedure to ensure comparability:

\begin{itemize}
    \item \textbf{Pre-training.} Pre-training (either self-supervised or through knowledge distillation) was performed on the synthetic transfer dataset. We used the AdamW optimizer with a learning rate schedule that included linear warm-up followed by cosine decay. Pre-training was conducted for 1,600 epochs.
    \item \textbf{Fine-tuning.} After pre-training, Vit tiny models were fine-tuned on the labeled portion of the target dataset using three different data scales. 
\end{itemize}

\subsubsection{Pre-Training Details}

Our pre-training setup follows the established practices for MAE and MoCo-v3, with certain adjustments tailored to the lightweight ViT-Tiny model. Key details regarding batch sizes and other parameters are described below.

For MAE pre-training, we used a batch size of 64. The training was conducted with the AdamW optimizer, starting with a base learning rate of $1.5 \times 10^{-4}$ and a weight decay of 0.05. The learning rate followed a cosine decay schedule, with the pre-training carried out for 1600 epochs. This configuration is based on the setup recommended~\cite{wang2023closer}.

For MoCo-v3 pre-training, we adopted a batch size of 32, again using the AdamW optimizer but with a slightly higher weight decay of 0.1. The base learning rate was set at $1.5 \times 10^{-4}$, with a momentum coefficient of 0.99 and a temperature parameter of 0.2. Similar to MAE, a cosine learning rate schedule was used, and the pre-training was run for 1600 epochs, following the protocol described~\cite{wang2023closer}.

In the Knowledge Distillation (KD) pre-training, we utilized a batch size of 48. The student model was trained using the Adam optimizer with a base learning rate of $1.5 \times 10^{-4}$ and a weight decay of 0.05. The learning rate followed a cosine decay schedule, and training was conducted for 1600 epochs to ensure sufficient convergence. This setup allowed us to effectively distill the knowledge from the SAM model into the smaller ViT-Tiny model for medical segmentation tasks.

All pre-training procedures were run on two NVIDIA L40 (48G) GPUs, which provided ample memory and computational power to handle the large-scale data and long training epochs efficiently.

\subsubsection{ViT-Tiny Fine-Tuning Details}

The ViT-Tiny model was fine-tuned using 160 epochs, with a batch size of 6 per GPU, and training distributed across 2 GPUs. A base learning rate of $1 \times 10^{-4}$ was applied and the AdamW optimizer was selected for fine-tuning. For reproducibility, deterministic training was enabled, and a random seed of $1234$ was set. Additionally, the loss function was a combination of cross-entropy loss and Dice loss, with a weighting of 0.2 for cross-entropy and 0.8 for Dice. 

\subsubsection{Evaluation Metrics}

We employed the following metrics to evaluate segmentation performance:

\begin{itemize}
    \item \textbf{Dice Similarity Coefficient (DSC).} Measures the overlap between the predicted and ground truth segmentation masks:
    \begin{equation}
    DSC = \frac{2 |A \cap B|}{|A| + |B|},
    \end{equation}
    where $A$ and $B$ represent the predicted and ground truth segmentation regions, respectively.
    
    \item \textbf{Hausdorff Distance 95 (HD95).} Measures the distance between the predicted and reference boundaries, providing insight into the model's ability to delineate structures accurately:
    \begin{equation}
    HD_{95} = \max \{ \underset{a \in A}{\operatorname{sup}} \min_{b \in B} d(a, b), \underset{b \in B}{\operatorname{sup}} \min_{a \in A} d(a, b) \},
    \end{equation}
    where $d(a, b)$ is the Euclidean distance between points $a$ and $b$.
    
    \item \textbf{Mean Intersection over Union (mIoU).} Quantifies the agreement between predicted and reference segmentation regions:
    \begin{equation}
    mIoU = \frac{1}{C} \sum_{c=1}^C \frac{|A_c \cap B_c|}{|A_c \cup B_c|},
    \end{equation}
    where $C$ represents the number of classes, and $A_c$ and $B_c$ are the predicted and ground truth regions for class $c$.
\end{itemize}

\subsection{Effectiveness of Task-specific Knowledge Distillation} 

\subsubsection{Performance Improvement}

\begin{table*}[!ht]
\setlength{\abovecaptionskip}{0pt}
\setlength{\belowcaptionskip}{0pt}
\centering
\large
\caption{Segmentation performance comparison for KidneyUS dataset across different methods, label counts, and transfer strategies}
\label{taba1}
\scalebox{0.36}{
\begin{tabular}{ccc
S[round-precision=4,round-mode=places]
S[round-precision=4,round-mode=places]
S[round-precision=4,round-mode=places]
S[round-precision=4,round-mode=places]
S[round-precision=4,round-mode=places]
S[round-precision=4,round-mode=places]
S[round-precision=4,round-mode=places]
S[round-precision=4,round-mode=places]
S[round-precision=4,round-mode=places]
S[round-precision=4,round-mode=places]}
\hline
\multirow{2}{*}{\textbf{Method Description}}  & \multirow{2}{*}{\textbf{Transfer Data Scale}} & \multirow{2}{*}{\textbf{Number of Labels}} & \multicolumn{2}{c}{\textbf{Capsule}} & \multicolumn{2}{c}{\textbf{Medulla}} & \multicolumn{2}{c}{\textbf{Central Echo Complex}} & \multicolumn{2}{c}{\textbf{Cortex}} & \multicolumn{2}{c}{\textbf{Mean}}      \\ 
                                              &                                               &                                            & \textbf{Dice}  & \textbf{HD95 (mm)}  & \textbf{Dice}  & \textbf{HD95 (mm)}  & \textbf{Dice}         & \textbf{HD95 (mm)}        & \textbf{Dice}  & \textbf{HD95 (mm)} & \textbf{Dice}     & \textbf{HD95 (mm)} \\ \hline
Small ViT Model Initialized Randomly          & -                                             & 80                                         & 0.365527       & 7.209279            & 0.563153       & 10.334204           & 0.497893              & 10.874268                 & 0.284101       & 19.775753          & 0.427669          & 12.048376          \\
Small ViT Model Initialized Randomly          & -                                             & 160                                        & 0.432659       & 6.746115            & 0.636476       & 8.098209            & 0.535125              & 15.370979                 & 0.380862       & 34.569200          & 0.496280          & 16.196126          \\
\rowcolor{lightblue}
Small ViT Model Initialized Randomly          & -                                             & 240                                        & 0.461553       & 7.958251            & 0.700183       & 6.795767            & 0.565286              & 20.472541                 & 0.434715       & 36.147944          & 0.540434          & 17.843626          \\ \hline
ImageNet Pre-trained Small ViT Model with MAE & -                                             & 80                                         & 0.366540       & 6.240602            & 0.510388       & 11.747848           & 0.459697              & 14.605546                 & 0.299946       & 25.229756          & 0.409143          & 14.455938          \\
ImageNet Pre-trained Small ViT Model with MAE & -                                             & 160                                        & 0.413429       & 7.357463            & 0.656852       & 7.172244            & 0.516009              & 20.364616                 & 0.362613       & 27.014086          & 0.487226          & 15.477102          \\
\rowcolor{lightblue}
ImageNet Pre-trained Small ViT Model with MAE & -                                             & 234                                        & 0.476499       & 7.446733            & 0.713396       & 5.706879            & 0.591659              & 19.022679                 & 0.398026       & 25.423874          & 0.544895 & 14.400041          \\ \hline
Self-supervised Pretraining with MoCo v3      & 1000                                          & 80                                         & 0.412357       & 6.889805            & 0.636843       & 7.056882            & 0.523286              & 10.231864                 & 0.388490       & 25.341662          & 0.490244          & 12.380053          \\
Self-supervised Pretraining with MoCo v3      & 1000                                          & 160                                        & 0.442654       & 8.060576            & 0.683864       & 6.576646            & 0.534931              & 15.980053                 & 0.427927       & 33.703756          & 0.522344          & 16.080258          \\
Self-supervised Pretraining with MoCo v3      & 1000                                          & 234                                        & 0.464931       & 7.707725            & 0.703281       & 8.048178            & 0.573090              & 17.900975                 & 0.449148       & 34.074437          & 0.547613          & 16.932829          \\
Self-supervised Pretraining with MoCo v3      & 2000                                          & 80                                         & 0.372451       & 8.405540            & 0.614426       & 10.733502           & 0.515452              & 11.621952                 & 0.395229       & 21.906503          & 0.474390          & 13.166874          \\
Self-supervised Pretraining with MoCo v3      & 2000                                          & 160                                        & 0.451473       & 10.364521           & 0.681220       & 7.962000            & 0.542801              & 20.746178                 & 0.430279       & 31.146802          & 0.526443          & 17.554875          \\
Self-supervised Pretraining with MoCo v3      & 2000                                          & 234                                        & 0.483436       & 6.654880            & 0.710756       & 5.595889            & 0.564536              & 25.767424                 & 0.458207       & 41.855285          & 0.554234          & 19.968370          \\
Self-supervised Pretraining with MoCo v3      & 3000                                          & 80                                         & 0.429286       & 6.261164            & 0.646649       & 6.644472            & 0.509476              & 11.963599                 & 0.390098       & 25.409734          & 0.493877          & 12.569742          \\
Self-supervised Pretraining with MoCo v3      & 3000                                          & 160                                        & 0.465552       & 7.172828            & 0.693699       & 6.989944            & 0.557788              & 15.291583                 & 0.452127       & 26.614794          & 0.542291          & 14.017287          \\
\rowcolor{lightblue}
Self-supervised Pretraining with MoCo v3      & 3000                                          & 234                                        & 0.478328       & 6.715730            & 0.698888       & 6.772106            & 0.560433              & 15.895159                 & 0.454711       & 35.576048          & 0.548090 & 16.239761          \\ \hline
Self-supervised Pretraining with MAE          & 1000                                          & 80                                         & 0.391319       & 7.143082            & 0.636498       & 7.681989            & 0.512799              & 10.991791                 & 0.391528       & 23.219233          & 0.483036          & 12.259024          \\
Self-supervised Pretraining with MAE          & 1000                                          & 160                                        & 0.489664       & 7.153026            & 0.716184       & 5.148957            & 0.567816              & 13.130932                 & 0.446481       & 29.667169          & 0.555036          & 13.775021          \\
Self-supervised Pretraining with MAE          & 1000                                          & 234                                        & 0.505742       & 5.607364            & 0.705290       & 5.059843            & 0.594797              & 11.960759                 & 0.490346       & 28.418441          & 0.574044          & 12.761602          \\
Self-supervised Pretraining with MAE          & 2000                                          & 80                                         & 0.455984       & 5.667931            & 0.682034       & 5.216553            & 0.543475              & 11.766397                 & 0.418547       & 24.136515          & 0.525010          & 11.696849          \\
Self-supervised Pretraining with MAE          & 2000                                          & 160                                        & 0.507894       & 6.269046            & 0.729567       & 5.070012            & 0.592112              & 14.439092                 & 0.501887       & 26.625965          & 0.582865          & 13.101029          \\
Self-supervised Pretraining with MAE          & 2000                                          & 234                                        & 0.512055       & 5.327603            & 0.727599       & 6.222383            & 0.607493              & 18.641464                 & 0.479976       & 36.988728          & 0.581781          & 16.795044          \\
Self-supervised Pretraining with MAE          & 3000                                          & 80                                         & 0.432937       & 5.861624            & 0.688392       & 4.908797            & 0.543139              & 12.123995                 & 0.421242       & 30.646326          & 0.521428          & 13.385186          \\
Self-supervised Pretraining with MAE          & 3000                                          & 160                                        & 0.472986       & 7.359776            & 0.715204       & 5.646196            & 0.567368              & 14.141013                 & 0.451613       & 26.535593          & 0.551793          & 13.420645          \\
\rowcolor{lightblue}
Self-supervised Pretraining with MAE          & 3000                                          & 234                                        & 0.508826       & 6.325348            & 0.727484       & 5.719492            & 0.610679              & 22.209104                 & 0.494853       & 37.039339          & 0.585460 & 17.823321          \\ \hline
Task-agnostic KD (SAM)                        & 1000                                          & 80                                         & 0.354395       & 15.193057           & 0.555822       & 8.282753            & 0.452854              & 13.300960                 & 0.287907       & 19.722102          & 0.412744          & 14.124718          \\
Task-agnostic KD (SAM)                        & 1000                                          & 160                                        & 0.452923       & 5.397341            & 0.680577       & 5.286007            & 0.562217              & 11.739660                 & 0.427083       & 17.308507          & 0.530700          & 9.932879           \\
Task-agnostic KD (SAM)                        & 1000                                          & 234                                        & 0.479127       & 6.392445            & 0.713356       & 5.935471            & 0.607868              & 14.313895                 & 0.475605       & 18.879178          & 0.568989          & 11.380247          \\
Task-agnostic KD (SAM)                        & 2000                                          & 80                                         & 0.393588       & 6.998950            & 0.571140       & 6.388580            & 0.489758              & 7.902897                  & 0.362388       & 21.175942          & 0.454218          & 10.616592          \\
Task-agnostic KD (SAM)                        & 2000                                          & 160                                        & 0.421718       & 8.242394            & 0.677754       & 6.911747            & 0.569816              & 9.399653                  & 0.437381       & 15.261986          & 0.526667          & 9.953945           \\
Task-agnostic KD (SAM)                        & 2000                                          & 234                                        & 0.454349       & 7.676564            & 0.717223       & 6.363116            & 0.612714              & 14.285555                 & 0.467273       & 24.772906          & 0.562890          & 13.274535          \\
Task-agnostic KD (SAM)                        & 3000                                          & 80                                         & 0.345722       & 10.149923           & 0.587619       & 8.991192            & 0.490568              & 14.387610                 & 0.333974       & 20.754578          & 0.439471          & 13.570826          \\
Task-agnostic KD (SAM)                        & 3000                                          & 160                                        & 0.463367       & 7.319346            & 0.687708       & 5.706525            & 0.557922              & 9.831575                  & 0.432118       & 22.794393          & 0.535279          & 11.412960          \\
\rowcolor{lightblue}
Task-agnostic KD (SAM)                        & 3000                                          & 234                                        & 0.466919       & 8.020589            & 0.710382       & 5.099528            & 0.583923              & 15.513128                 & 0.469477       & 28.416192          & 0.557675 & 14.262359          \\ \hline
Task-specific KD (SAM)                        & 1000                                          & 80                                         & 0.404172       & 8.655597            & 0.628322       & 8.094922            & 0.539420              & 12.367897                 & 0.420227       & 21.267318          & 0.498035          & 12.596433          \\
Task-specific KD (SAM)                        & 1000                                          & 160                                        & 0.486351       & 6.320157            & 0.703741       & 5.805178            & 0.590818              & 12.301863                 & 0.465022       & 24.049936          & 0.561483          & 12.119283          \\
Task-specific KD (SAM)                        & 1000                                          & 234                                        & 0.504028       & 6.486578            & 0.731509       & 5.088478            & 0.621117              & 10.145889                 & 0.478092       & 25.226154          & 0.583686          & 11.736775          \\
Task-specific KD (SAM)                        & 2000                                          & 80                                         & 0.424325       & 7.411348            & 0.685504       & 6.801277            & 0.597434              & 11.395934                 & 0.476650       & 21.041810          & 0.545978          & 11.662592          \\
Task-specific KD (SAM)                        & 2000                                          & 160                                        & 0.489206       & 10.860590           & 0.751264       & 5.069378            & 0.628475              & 10.000711                 & 0.519026       & 25.553433          & 0.596993          & 12.871028          \\
Task-specific KD (SAM)                        & 2000                                          & 234                                        & 0.522575       & 6.166045            & 0.757496       & 4.890252            & 0.626832              & 13.678913                 & 0.511779       & 33.856540          & 0.604670          & 14.647937          \\
Task-specific KD (SAM)                        & 3000                                          & 80                                         & 0.451852       & 7.377404            & 0.719782       & 5.749098            & 0.604871              & 12.721272                 & 0.469850       & 28.571851          & 0.561589          & 13.604906          \\
Task-specific KD (SAM)                        & 3000                                          & 160                                        & 0.520102       & 6.842670            & 0.763957       & 5.091122            & 0.650946              & 10.012059                 & 0.525993       & 25.010801          & 0.615249          & 11.739163          \\
\rowcolor{lightblue}
Task-specific KD (SAM)                        & 3000                                          & 234                                        & 0.539609       & 6.054890            & 0.761477       & 5.670974            & 0.630104              & 17.205964                 & 0.524243       & 37.626323          & 0.613858 & 16.639538          \\ 
\hline
\end{tabular}
}
\vspace{-.1in}
\end{table*}

\begin{table*}[!ht]
\setlength{\abovecaptionskip}{0pt}
\setlength{\belowcaptionskip}{0pt}
\centering
\large
\caption{Segmentation performance comparison for Autooral dataset across different methods, label counts, and transfer strategies.}
\scalebox{0.42}{
\begin{tabular}{ccc
S[round-precision=4,round-mode=places]
S[round-precision=4,round-mode=places]
S[round-precision=4,round-mode=places]}
\hline
\textbf{Method Description}                   & \textbf{Transfer Data Scale} & \textbf{Number of Labels} & \textbf{Dice}     & \textbf{HD95 (mm)} & \textbf{mIou} \\ \hline
Small ViT Model Initialized Randomly          & -                            & 100                       & 0.343735          & 31.769134          & 0.209871      \\
Small ViT Model Initialized Randomly          & -                            & 200                       & 0.468099          & 13.578084          & 0.307528      \\
\rowcolor{lightblue}
Small ViT Model Initialized Randomly          & -                            & 300                       & 0.447014 & 18.451650          & 0.290578      \\ \hline
ImageNet Pre-trained Small ViT Model with MAE & -                            & 100                       & 0.392410          & 19.156974          & 0.247065      \\
ImageNet Pre-trained Small ViT Model with MAE & -                            & 200                       & 0.545745          & 10.047031          & 0.377533      \\
\rowcolor{lightblue}
ImageNet Pre-trained Small ViT Model with MAE & -                            & 300                       & 0.549543 & 12.709526          & 0.385966      \\ \hline
Self-supervised Pretraining with MoCo v3      & 1000                         & 100                       & 0.345393          & 28.507897          & 0.214660      \\
Self-supervised Pretraining with MoCo v3      & 1000                         & 200                       & 0.478416          & 12.118136          & 0.318253      \\
Self-supervised Pretraining with MoCo v3      & 1000                         & 300                       & 0.483816          & 15.758545          & 0.321433      \\
Self-supervised Pretraining with MoCo v3      & 2000                         & 100                       & 0.354799          & 26.403356          & 0.222320      \\
Self-supervised Pretraining with MoCo v3      & 2000                         & 200                       & 0.450086          & 13.880922          & 0.295264      \\
Self-supervised Pretraining with MoCo v3      & 2000                         & 300                       & 0.524400          & 12.634776          & 0.357497      \\
Self-supervised Pretraining with MoCo v3      & 3000                         & 100                       & 0.356892          & 26.867755          & 0.222765      \\
Self-supervised Pretraining with MoCo v3      & 3000                         & 200                       & 0.506053          & 11.189565          & 0.341772      \\
\rowcolor{lightblue}
Self-supervised Pretraining with MoCo v3      & 3000                         & 300                       & 0.463555 & 16.725683          & 0.304908      \\ \hline
Self-supervised Pretraining with MAE          & 1000                         & 100                       & 0.367109          & 23.279609          & 0.228835      \\
Self-supervised Pretraining with MAE          & 1000                         & 200                       & 0.507393          & 10.262630          & 0.341763      \\
Self-supervised Pretraining with MAE          & 1000                         & 300                       & 0.483301          & 14.876285          & 0.323149      \\
Self-supervised Pretraining with MAE          & 2000                         & 100                       & 0.346559          & 23.414483          & 0.213461      \\
Self-supervised Pretraining with MAE          & 2000                         & 200                       & 0.501595          & 11.448577          & 0.337278      \\
Self-supervised Pretraining with MAE          & 2000                         & 300                       & 0.506041          & 14.123659          & 0.342482      \\
Self-supervised Pretraining with MAE          & 3000                         & 100                       & 0.353016          & 23.872370          & 0.218043      \\
Self-supervised Pretraining with MAE          & 3000                         & 200                       & 0.511242          & 11.794915          & 0.345717      \\
\rowcolor{lightblue}
Self-supervised Pretraining with MAE          & 3000                         & 300                       & 0.541607 & 14.165245          & 0.375148      \\ \hline
Task-agnostic KD (SAM)                        & 1000                         & 100                       & 0.311157          & 39.661063          & 0.187392      \\
Task-agnostic KD (SAM)                        & 1000                         & 200                       & 0.529457          & 11.131883          & 0.362852      \\
Task-agnostic KD (SAM)                        & 1000                         & 300                       & 0.511617          & 11.831350          & 0.346203      \\
Task-agnostic KD (SAM)                        & 2000                         & 100                       & 0.270911          & 42.610686          & 0.157511      \\
Task-agnostic KD (SAM)                        & 2000                         & 200                       & 0.544727          & 9.844783           & 0.375239      \\
Task-agnostic KD (SAM)                        & 2000                         & 300                       & 0.556253          & 14.850533          & 0.389058      \\
Task-agnostic KD (SAM)                        & 3000                         & 100                       & 0.390172          & 22.135001          & 0.244522      \\
Task-agnostic KD (SAM)                        & 3000                         & 200                       & 0.510872          & 10.419124          & 0.343232      \\
\rowcolor{lightblue}
Task-agnostic KD (SAM)                        & 3000                         & 300                       & 0.568807 & 13.221038          & 0.401103      \\ \hline
Task-specific KD (SAM)                        & 1000                         & 100                       & 0.270951          & 44.668097          & 0.157677      \\
Task-specific KD (SAM)                        & 1000                         & 200                       & 0.517554          & 10.018510          & 0.351274      \\
Task-specific KD (SAM)                        & 1000                         & 300                       & 0.532011          & 16.264691          & 0.365525      \\
Task-specific KD (SAM)                        & 2000                         & 100                       & 0.399653          & 27.869315          & 0.255415      \\
Task-specific KD (SAM)                        & 2000                         & 200                       & 0.530774          & 13.706525          & 0.361736      \\
Task-specific KD (SAM)                        & 2000                         & 300                       & 0.573868          & 9.061819           & 0.405419      \\
Task-specific KD (SAM)                        & 3000                         & 100                       & 0.384006          & 24.861660          & 0.241281      \\
Task-specific KD (SAM)                        & 3000                         & 200                       & 0.557013          & 12.394325          & 0.387265      \\
\rowcolor{lightblue}
Task-specific KD (SAM)                        & 3000                         & 300                       & 0.575414 & 9.572240           & 0.407842      \\ \hline
\label{taba2}
\end{tabular}}
\vspace{-.1in}
\end{table*}

\begin{table*}[!ht]
\setlength{\abovecaptionskip}{0pt}
\setlength{\belowcaptionskip}{0pt}
\centering
% \small
% \normalsize
\large
\caption{Segmentation performance comparison for CHAOS dataset across different methods, label counts, and transfer strategies.}
\scalebox{0.37}{
\begin{tabular}{ccc
S[round-precision=4,round-mode=places]
S[round-precision=4,round-mode=places]
S[round-precision=4,round-mode=places]
S[round-precision=4,round-mode=places]
S[round-precision=4,round-mode=places]
S[round-precision=4,round-mode=places]
S[round-precision=4,round-mode=places]
S[round-precision=4,round-mode=places]
S[round-precision=4,round-mode=places]
S[round-precision=4,round-mode=places]}
\hline
\multirow{2}{*}{\textbf{Method Description}}  & \multirow{2}{*}{\textbf{Transfer Data Scale}} & \multirow{2}{*}{\textbf{Number of Labels}} & \multicolumn{2}{c}{\textbf{Liver}} & \multicolumn{2}{c}{\textbf{Right kidney}} & \multicolumn{2}{c}{\textbf{Left kidney}} & \multicolumn{2}{c}{\textbf{Spleen}} & \multicolumn{2}{c}{\textbf{Mean}}      \\  
                                              &                                               &                                            & \textbf{Dice} & \textbf{HD95 (mm)} & \textbf{Dice}     & \textbf{HD95 (mm)}    & \textbf{Dice}    & \textbf{HD95 (mm)}    & \textbf{Dice}  & \textbf{HD95 (mm)} & \textbf{Dice}     & \textbf{HD95 (mm)} \\ \hline
                                              \rowcolor{lightblue}
Small ViT Model Initialized Randomly          & -                                             & 100                                        & 0.769331      & 5.576691           & 0.560791          & 13.841846             & 0.494679         & 17.183394             & 0.503031       & 9.139512           & 0.581958          & 11.435361          \\
Small ViT Model Initialized Randomly          & -                                             & 200                                        & 0.836841      & 3.170339           & 0.722532          & 8.874210              & 0.669306         & 12.153458             & 0.597167       & 9.159796           & 0.706462          & 8.339451           \\
Small ViT Model Initialized Randomly          & -                                             & 296                                        & 0.860521      & 2.750788           & 0.811055          & 12.238151             & 0.796047         & 4.384841              & 0.777899       & 3.368866           & 0.811380 & 5.685661           \\ \hline
\rowcolor{lightblue}
ImageNet Pre-trained Small ViT Model with MAE & -                                             & 100                                        & 0.791981      & 4.266231           & 0.595911          & 21.812918             & 0.582712         & 25.112642             & 0.580629       & 18.205838          & 0.637808          & 17.349407          \\
ImageNet Pre-trained Small ViT Model with MAE & -                                             & 200                                        & 0.869700      & 2.537169           & 0.795160          & 5.332106              & 0.778549         & 8.374767              & 0.693145       & 9.150289           & 0.784138          & 6.348583           \\
ImageNet Pre-trained Small ViT Model with MAE & -                                             & 296                                        & 0.903310      & 1.848208           & 0.846232          & 18.894327             & 0.845360         & 1.896025              & 0.838013       & 1.744327           & 0.858229 & 6.095722           \\ \hline
Self-supervised Pretraining with MoCo v3      & 1000                                          & 100                                        & 0.777349      & 6.055418           & 0.646548          & 13.447892             & 0.609572         & 10.956909             & 0.608271       & 7.783144           & 0.660435          & 9.560841           \\
Self-supervised Pretraining with MoCo v3      & 1000                                          & 200                                        & 0.862343      & 2.507813           & 0.793501          & 12.667774             & 0.773489         & 3.550156              & 0.674011       & 4.677557           & 0.775836          & 5.850825           \\
Self-supervised Pretraining with MoCo v3      & 1000                                          & 296                                        & 0.889730      & 1.952850           & 0.845787          & 13.499914             & 0.861738         & 1.634371              & 0.827290       & 1.547524           & 0.856136          & 4.658665           \\
Self-supervised Pretraining with MoCo v3      & 2000                                          & 100                                        & 0.783017      & 5.565168           & 0.632462          & 14.341788             & 0.587185         & 7.477205              & 0.579076       & 8.138245           & 0.645435          & 8.880601           \\
Self-supervised Pretraining with MoCo v3      & 2000                                          & 200                                        & 0.859786      & 2.539581           & 0.804344          & 11.690187             & 0.778046         & 3.827255              & 0.708477       & 4.507085           & 0.787663          & 5.641027           \\
Self-supervised Pretraining with MoCo v3      & 2000                                          & 296                                        & 0.879381      & 2.738523           & 0.829614          & 11.051095             & 0.823712         & 2.896150              & 0.840594       & 1.595434           & 0.843325          & 4.570300           \\
\rowcolor{lightblue}
Self-supervised Pretraining with MoCo v3      & 3000                                          & 100                                        & 0.784902      & 6.355750           & 0.575882          & 9.180425              & 0.574123         & 20.950770             & 0.601032       & 15.595047          & 0.633985          & 13.020498          \\
Self-supervised Pretraining with MoCo v3      & 3000                                          & 200                                        & 0.865341      & 2.230002           & 0.810211          & 9.353493              & 0.763350         & 3.999258              & 0.695023       & 5.169722           & 0.783481          & 5.188119           \\
Self-supervised Pretraining with MoCo v3      & 3000                                          & 296                                        & 0.888298      & 1.882564           & 0.835035          & 5.706071              & 0.843603         & 1.951167              & 0.832278       & 1.831569           & 0.849804 & 2.842843           \\ \hline
Self-supervised Pretraining with MAE          & 1000                                          & 100                                        & 0.752549      & 5.786721           & 0.612039          & 19.060865             & 0.572355         & 8.441048              & 0.550281       & 8.392658           & 0.621806          & 10.420323          \\
Self-supervised Pretraining with MAE          & 1000                                          & 200                                        & 0.867528      & 2.120173           & 0.805714          & 9.542262              & 0.756770         & 4.268670              & 0.656325       & 6.291894           & 0.771584          & 5.555750           \\
Self-supervised Pretraining with MAE          & 1000                                          & 296                                        & 0.896837      & 1.564614           & 0.851349          & 12.504035             & 0.868408         & 1.536318              & 0.821839       & 1.557181           & 0.859608          & 4.290537           \\
Self-supervised Pretraining with MAE          & 2000                                          & 100                                        & 0.771572      & 5.469409           & 0.628462          & 17.205783             & 0.570193         & 8.427786              & 0.562606       & 8.760171           & 0.633208          & 9.965787           \\
Self-supervised Pretraining with MAE          & 2000                                          & 200                                        & 0.866160      & 2.421995           & 0.815123          & 5.854077              & 0.761877         & 3.659853              & 0.650437       & 6.003624           & 0.773399          & 4.484887           \\
Self-supervised Pretraining with MAE          & 2000                                          & 296                                        & 0.892222      & 1.696078           & 0.843554          & 4.300517              & 0.851209         & 1.943091              & 0.817314       & 1.814975           & 0.851075          & 2.438665           \\
\rowcolor{lightblue}
Self-supervised Pretraining with MAE          & 3000                                          & 100                                        & 0.773710      & 5.113235           & 0.607173          & 13.843094             & 0.587308         & 12.829990             & 0.521663       & 8.891247           & 0.622464          & 10.169391          \\
Self-supervised Pretraining with MAE          & 3000                                          & 200                                        & 0.864448      & 2.321031           & 0.815122          & 9.463399              & 0.759516         & 3.986791              & 0.684159       & 4.632103           & 0.780811          & 5.100831           \\
Self-supervised Pretraining with MAE          & 3000                                          & 296                                        & 0.894894      & 1.610434           & 0.859902          & 5.936901              & 0.859551         & 1.766207              & 0.831951       & 2.247190           & 0.861574 & 2.890183           \\ \hline
Task-agnostic KD (SAM)                        & 1000                                          & 100                                        & 0.770699      & 5.834436           & 0.557025          & 17.527080             & 0.538105         & 15.019709             & 0.513968       & 8.559629           & 0.594949          & 11.735213          \\
Task-agnostic KD (SAM)                        & 1000                                          & 200                                        & 0.854134      & 2.934486           & 0.787699          & 6.950461              & 0.712642         & 4.822350              & 0.563576       & 7.767248           & 0.729513          & 5.618636           \\
Task-agnostic KD (SAM)                        & 1000                                          & 296                                        & 0.865981      & 2.949987           & 0.822227          & 11.814762             & 0.776008         & 6.487862              & 0.772265       & 3.249106           & 0.809120          & 6.125429           \\
Task-agnostic KD (SAM)                        & 2000                                          & 100                                        & 0.781638      & 5.608377           & 0.652019          & 8.016250              & 0.577601         & 6.523103              & 0.598655       & 10.138590          & 0.652478          & 7.571580           \\
Task-agnostic KD (SAM)                        & 2000                                          & 200                                        & 0.854857      & 2.509578           & 0.784865          & 4.606088              & 0.736324         & 4.635503              & 0.709748       & 4.927099           & 0.771448          & 4.169567           \\
Task-agnostic KD (SAM)                        & 2000                                          & 296                                        & 0.890072      & 1.776057           & 0.828861          & 9.649415              & 0.787322         & 6.007818              & 0.804283       & 3.040209           & 0.827635          & 5.118375           \\
\rowcolor{lightblue}
Task-agnostic KD (SAM)                        & 3000                                          & 100                                        & 0.814777      & 3.540219           & 0.608326          & 6.915636              & 0.645111         & 6.996941              & 0.609132       & 8.591247           & 0.669336          & 6.511011           \\
Task-agnostic KD (SAM)                        & 3000                                          & 200                                        & 0.861439      & 2.385307           & 0.785372          & 4.340291              & 0.716854         & 10.882333             & 0.720232       & 4.024512           & 0.770974          & 5.408111           \\
Task-agnostic KD (SAM)                        & 3000                                          & 296                                        & 0.893529      & 1.965452           & 0.867969          & 9.210443              & 0.838220         & 2.496637              & 0.786236       & 4.459842           & 0.846488 & 4.533094           \\ \hline
Task-specific KD (SAM)                        & 1000                                          & 100                                        & 0.714310      & 8.998159           & 0.455175          & 24.111425             & 0.471007         & 11.793728             & 0.479716       & 13.232267          & 0.530052          & 14.533895          \\
Task-specific KD (SAM)                        & 1000                                          & 200                                        & 0.852681      & 2.863821           & 0.801812          & 17.042554             & 0.749188         & 4.288404              & 0.641607       & 6.647172           & 0.761322          & 7.710488           \\
Task-specific KD (SAM)                        & 1000                                          & 296                                        & 0.875601      & 2.249210           & 0.837024          & 17.134803             & 0.834653         & 1.882431              & 0.752044       & 4.353539           & 0.824831          & 6.404996           \\
Task-specific KD (SAM)                        & 2000                                          & 100                                        & 0.806151      & 7.908159           & 0.667481          & 23.736149             & 0.668858         & 8.383064              & 0.551066       & 8.914340           & 0.673389          & 12.235428          \\
Task-specific KD (SAM)                        & 2000                                          & 200                                        & 0.864673      & 2.353434           & 0.812816          & 8.830353              & 0.794810         & 3.473729              & 0.712618       & 4.710153           & 0.796229          & 4.841917           \\
Task-specific KD (SAM)                        & 2000                                          & 296                                        & 0.899287      & 1.815036           & 0.854299          & 14.452029             & 0.856734         & 1.633953              & 0.830011       & 1.620766           & 0.860083          & 4.880446           \\
\rowcolor{lightblue}
Task-specific KD (SAM)                        & 3000                                          & 100                                        & 0.818609      & 7.601213           & 0.658769          & 22.621407             & 0.657114         & 5.764730              & 0.630481       & 8.495202           & 0.691243          & 11.120638          \\
Task-specific KD (SAM)                        & 3000                                          & 200                                        & 0.875734      & 2.088300           & 0.825809          & 9.768370              & 0.801610         & 3.691354              & 0.706773       & 4.720077           & 0.802481          & 5.067025           \\
Task-specific KD (SAM)                        & 3000                                          & 296                                        & 0.895469      & 2.077859           & 0.855298          & 11.802779             & 0.867955         & 1.369500              & 0.846288       & 1.495408           & 0.866252 & 4.186386           \\ \hline
\label{taba3}
\end{tabular}}
\vspace{-.1in}
\end{table*}

\begin{table*}[!ht]
\setlength{\abovecaptionskip}{0pt}
\setlength{\belowcaptionskip}{0pt}
\centering
\large
\caption{Segmentation performance comparison for PH2 dataset across different methods, label counts, and transfer strategies.}
\scalebox{0.46}{
\begin{tabular}{ccc
S[round-precision=4,round-mode=places]
S[round-precision=4,round-mode=places]
S[round-precision=4,round-mode=places]}
\hline
\textbf{Method Description} & \textbf{Transfer Data Scale} & \textbf{Number of Labels} & \textbf{Dice}              & \textbf{HD95 (mm)} & \textbf{mIou}     \\ \hline
Small ViT Model Initialized Randomly          & -                   & 40               & 0.858609          & 6.360275  & 0.760908 \\
Small ViT Model Initialized Randomly          & -                   & 80               & 0.886165          & 3.000000  & 0.797038 \\
\rowcolor{lightblue}
Small ViT Model Initialized Randomly          & -                   & 160              & 0.917562 & 2.000000  & 0.848818 \\ \hline
ImageNet Pre-trained Small ViT Model with MAE & -                   & 40               & 0.898042          & 2.868034  & 0.816172 \\
ImageNet Pre-trained Small ViT Model with MAE & -                   & 80               & 0.907811          & 2.000000  & 0.832361 \\
\rowcolor{lightblue}
ImageNet Pre-trained Small ViT Model with MAE & -                   & 160              & 0.926610 & 1.559017  & 0.863431 \\ \hline
Self-supervised Pretraining with MoCo v3      & 1000                & 40               & 0.885069          & 2.750000  & 0.795992 \\
Self-supervised Pretraining with MoCo v3      & 1000                & 80               & 0.870842          & 4.750000  & 0.778971 \\
Self-supervised Pretraining with MoCo v3      & 1000                & 160              & 0.939136          & 1.500000  & 0.885451 \\
Self-supervised Pretraining with MoCo v3      & 2000                & 40               & 0.885710          & 3.000000  & 0.795910 \\
Self-supervised Pretraining with MoCo v3      & 2000                & 80               & 0.901938          & 3.326784  & 0.823245 \\
Self-supervised Pretraining with MoCo v3      & 2000                & 160              & 0.941540          & 1.207107  & 0.889793 \\
Self-supervised Pretraining with MoCo v3      & 3000                & 40               & 0.896513          & 2.309017  & 0.813311 \\
Self-supervised Pretraining with MoCo v3      & 3000                & 80               & 0.885384          & 2.579156  & 0.796889 \\
\rowcolor{lightblue}
Self-supervised Pretraining with MoCo v3      & 3000                & 160              & 0.935063 & 1.603553  & 0.878177 \\ \hline
Self-supervised Pretraining with MAE          & 1000                & 40               & 0.897060          & 2.309017  & 0.814387 \\
Self-supervised Pretraining with MAE          & 1000                & 80               & 0.902451          & 2.250000  & 0.823193 \\
Self-supervised Pretraining with MAE          & 1000                & 160              & 0.939100          & 1.457107  & 0.885571 \\
Self-supervised Pretraining with MAE          & 2000                & 40               & 0.892608          & 2.500000  & 0.807242 \\
Self-supervised Pretraining with MAE          & 2000                & 80               & 0.900187          & 2.250000  & 0.819649 \\
Self-supervised Pretraining with MAE          & 2000                & 160              & 0.938993          & 1.603553  & 0.885311 \\
Self-supervised Pretraining with MAE          & 3000                & 40               & 0.888935          & 2.559017  & 0.801529 \\
Self-supervised Pretraining with MAE          & 3000                & 80               & 0.899152          & 2.250000  & 0.818044 \\
\rowcolor{lightblue}
Self-supervised Pretraining with MAE          & 3000                & 160              & 0.940803 & 1.603553  & 0.888602 \\ \hline
Task-agnostic KD (SAM)                        & 1000                & 40               & 0.871900          & 4.000000  & 0.775168 \\
Task-agnostic KD (SAM)                        & 1000                & 80               & 0.877354          & 2.750000  & 0.783152 \\
Task-agnostic KD (SAM)                        & 1000                & 160              & 0.916291          & 2.000000  & 0.845768 \\
Task-agnostic KD (SAM)                        & 2000                & 40               & 0.850109          & 5.293245  & 0.744679 \\
Task-agnostic KD (SAM)                        & 2000                & 80               & 0.871551          & 3.927051  & 0.774184 \\
Task-agnostic KD (SAM)                        & 2000                & 160              & 0.907940          & 2.207107  & 0.831700 \\
Task-agnostic KD (SAM)                        & 3000                & 40               & 0.865926          & 3.809017  & 0.767337 \\
Task-agnostic KD (SAM)                        & 3000                & 80               & 0.887228          & 2.750000  & 0.798400 \\
\rowcolor{lightblue}
Task-agnostic KD (SAM)                        & 3000                & 160              & 0.901878 & 2.750000  & 0.821489 \\ \hline
Task-specific KD (SAM)                        & 1000                & 40               & 0.871118          & 4.020691  & 0.774603 \\
Task-specific KD (SAM)                        & 1000                & 80               & 0.890806          & 2.750000  & 0.804287 \\
Task-specific KD (SAM)                        & 1000                & 160              & 0.932858          & 1.500000  & 0.874377 \\
Task-specific KD (SAM)                        & 2000                & 40               & 0.865453          & 4.065601  & 0.765033 \\
Task-specific KD (SAM)                        & 2000                & 80               & 0.888767          & 3.000000  & 0.801247 \\
Task-specific KD (SAM)                        & 2000                & 160              & 0.937292          & 1.500000  & 0.882300 \\
Task-specific KD (SAM)                        & 3000                & 40               & 0.882629          & 3.191957  & 0.790852 \\
Task-specific KD (SAM)                        & 3000                & 80               & 0.907515          & 2.059017  & 0.831390 \\
\rowcolor{lightblue}
Task-specific KD (SAM)                        & 3000                & 160              & 0.945516 & 1.353553  & 0.897230 \\ \hline
\label{taba4}
\end{tabular}}
\vspace{-.1in}
\end{table*}

\begin{table*}[!ht]
\setlength{\abovecaptionskip}{0pt}
\setlength{\belowcaptionskip}{0pt}
\centering
% \small
% \normalsize
\large
\caption{Segmentation performance comparison for DRIVE dataset across different methods, label counts, and transfer strategies.}
\scalebox{0.48}{
\begin{tabular}{ccc
S[round-precision=4,round-mode=places]
S[round-precision=4,round-mode=places]
S[round-precision=4,round-mode=places]}
\hline
\textbf{Method Description}                   & \textbf{Transfer Data Scale} & \textbf{Number of Labels} & \textbf{Dice}     & \textbf{HD95 (mm)} & \textbf{mIou} \\ \hline
Small ViT Model Initialized Randomly          & -                            & 5                         & 0.216444          & 54.654844          & 0.121393      \\
Small ViT Model Initialized Randomly          & -                            & 10                        & 0.207636          & 55.547012          & 0.115857      \\
\rowcolor{lightblue}
Small ViT Model Initialized Randomly          & -                            & 20                        & 0.328391 & 45.043437          & 0.196459      \\ \hline
ImageNet Pre-trained Small ViT Model with MAE & -                            & 5                         & 0.283369          & 51.523861          & 0.165082      \\
ImageNet Pre-trained Small ViT Model with MAE & -                            & 10                        & 0.288355          & 52.926785          & 0.168485      \\
\rowcolor{lightblue}
ImageNet Pre-trained Small ViT Model with MAE & -                            & 20                        & 0.390915 & 44.078095          & 0.242950      \\ \hline
Self-supervised Pretraining with MoCo v3      & 1000                         & 5                         & 0.228309          & 54.242483          & 0.128927      \\
Self-supervised Pretraining with MoCo v3      & 1000                         & 10                        & 0.232960          & 54.867728          & 0.131854      \\
Self-supervised Pretraining with MoCo v3      & 1000                         & 20                        & 0.335231          & 45.239209          & 0.201376      \\
Self-supervised Pretraining with MoCo v3      & 2000                         & 5                         & 0.226954          & 54.360813          & 0.128063      \\
Self-supervised Pretraining with MoCo v3      & 2000                         & 10                        & 0.226894          & 55.059324          & 0.127979      \\
Self-supervised Pretraining with MoCo v3      & 2000                         & 20                        & 0.334053          & 45.283416          & 0.200526      \\
Self-supervised Pretraining with MoCo v3      & 3000                         & 5                         & 0.232731          & 53.886518          & 0.131745      \\
Self-supervised Pretraining with MoCo v3      & 3000                         & 10                        & 0.226114          & 55.081948          & 0.127483      \\
\rowcolor{lightblue}
Self-supervised Pretraining with MoCo v3      & 3000                         & 20                        & 0.337846 & 45.021219          & 0.203270      \\ \hline
Self-supervised Pretraining with MAE          & 1000                         & 5                         & 0.257815          & 51.762717          & 0.148010      \\
Self-supervised Pretraining with MAE          & 1000                         & 10                        & 0.243740          & 54.876674          & 0.138806      \\
Self-supervised Pretraining with MAE          & 1000                         & 20                        & 0.360722          & 44.553366          & 0.220058      \\
Self-supervised Pretraining with MAE          & 2000                         & 5                         & 0.262314          & 52.001732          & 0.150977      \\
Self-supervised Pretraining with MAE          & 2000                         & 10                        & 0.268145          & 53.965299          & 0.154864      \\
Self-supervised Pretraining with MAE          & 2000                         & 20                        & 0.367758          & 44.429429          & 0.225319      \\
Self-supervised Pretraining with MAE          & 3000                         & 5                         & 0.266487          & 51.529892          & 0.153740      \\
Self-supervised Pretraining with MAE          & 3000                         & 10                        & 0.272239          & 53.653587          & 0.157604      \\
\rowcolor{lightblue}
Self-supervised Pretraining with MAE          & 3000                         & 20                        & 0.368881 & 44.254429          & 0.226164      \\ \hline
Task-agnostic KD (SAM)                        & 1000                         & 5                         & 0.168898          & 23.267621          & 0.092277      \\
Task-agnostic KD (SAM)                        & 1000                         & 10                        & 0.213436          & 7.736254           & 0.119474      \\
Task-agnostic KD (SAM)                        & 1000                         & 20                        & 0.447215          & 5.608058           & 0.288010      \\
Task-agnostic KD (SAM)                        & 2000                         & 5                         & 0.154042          & 13.658658          & 0.083449      \\
Task-agnostic KD (SAM)                        & 2000                         & 10                        & 0.277675          & 9.325992           & 0.161246      \\
Task-agnostic KD (SAM)                        & 2000                         & 20                        & 0.409710          & 8.674235           & 0.257632      \\
Task-agnostic KD (SAM)                        & 3000                         & 5                         & 0.193542          & 8.405125           & 0.107148      \\
Task-agnostic KD (SAM)                        & 3000                         & 10                        & 0.232445          & 8.908811           & 0.131545      \\
\rowcolor{lightblue}
Task-agnostic KD (SAM)                        & 3000                         & 20                        & 0.414178 & 6.646933           & 0.261177      \\ \hline
Task-specific KD (SAM)                        & 1000                         & 5                         & 0.220737          & 5.590891           & 0.124093      \\
Task-specific KD (SAM)                        & 1000                         & 10                        & 0.303974          & 8.859618           & 0.179227      \\
Task-specific KD (SAM)                        & 1000                         & 20                        & 0.493976          & 5.470358           & 0.328003      \\
Task-specific KD (SAM)                        & 2000                         & 5                         & 0.237599          & 6.699697           & 0.134859      \\
Task-specific KD (SAM)                        & 2000                         & 10                        & 0.387303          & 5.238613           & 0.240167      \\
Task-specific KD (SAM)                        & 2000                         & 20                        & 0.504940          & 4.236068           & 0.337750      \\
Task-specific KD (SAM)                        & 3000                         & 5                         & 0.288590          & 4.345208           & 0.168681      \\
Task-specific KD (SAM)                        & 3000                         & 10                        & 0.435124          & 5.288123           & 0.278059      \\
\rowcolor{lightblue}
Task-specific KD (SAM)                        & 3000                         & 20                        & 0.574099 & 3.802776           & 0.402628      \\ \hline
\label{taba5}
\end{tabular}}
\vspace{-.1in}
\end{table*}

\begin{figure}[htbp]
    \centering
    \includegraphics[scale=0.42]{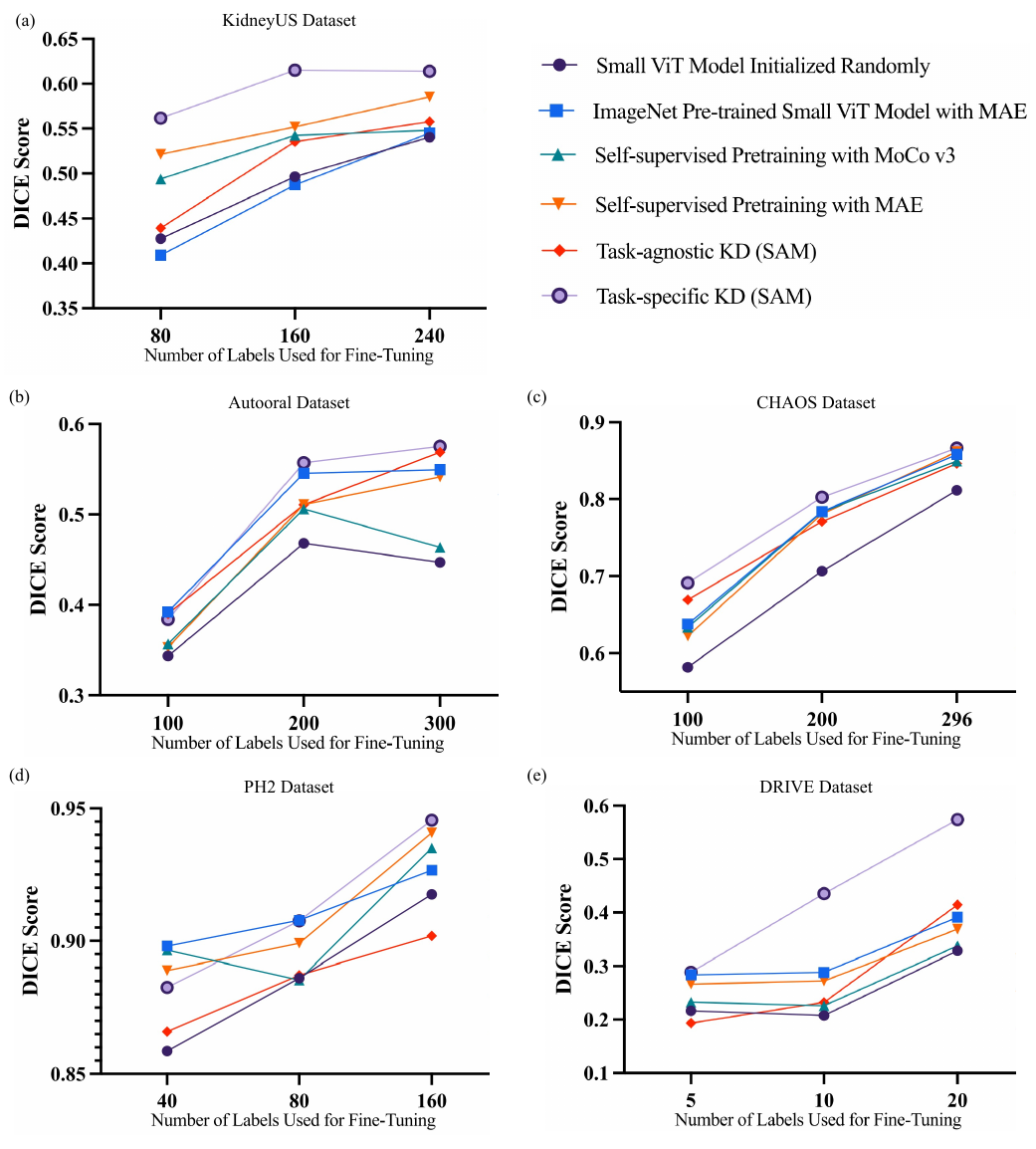}
    \caption{Comparison of Dice scores across five medical imaging datasets (KidneyUS, Autooral, CHAOS, PH2, and DRIVE) using different pretraining and knowledge distillation methods. Each plot (a)-(e) represents a dataset, with results shown for models fine-tuned with varying numbers of labeled samples. Approaches include a Small ViT model trained from scratch, ImageNet pre-trained Small ViT model with MAE, self-supervised pretraining with MoCo v3 and MAE, task-agnostic knowledge distillation (KD), and task-specific KD (TS-KD). Task-specific KD consistently demonstrates the highest performance trends and superior Dice scores across all datasets, showcasing its effectiveness in leveraging task-relevant features for medical image segmentation.}
    \label{fig10}
\end{figure}

\begin{figure*}[htbp]
    \centering
    \includegraphics[scale=0.79]{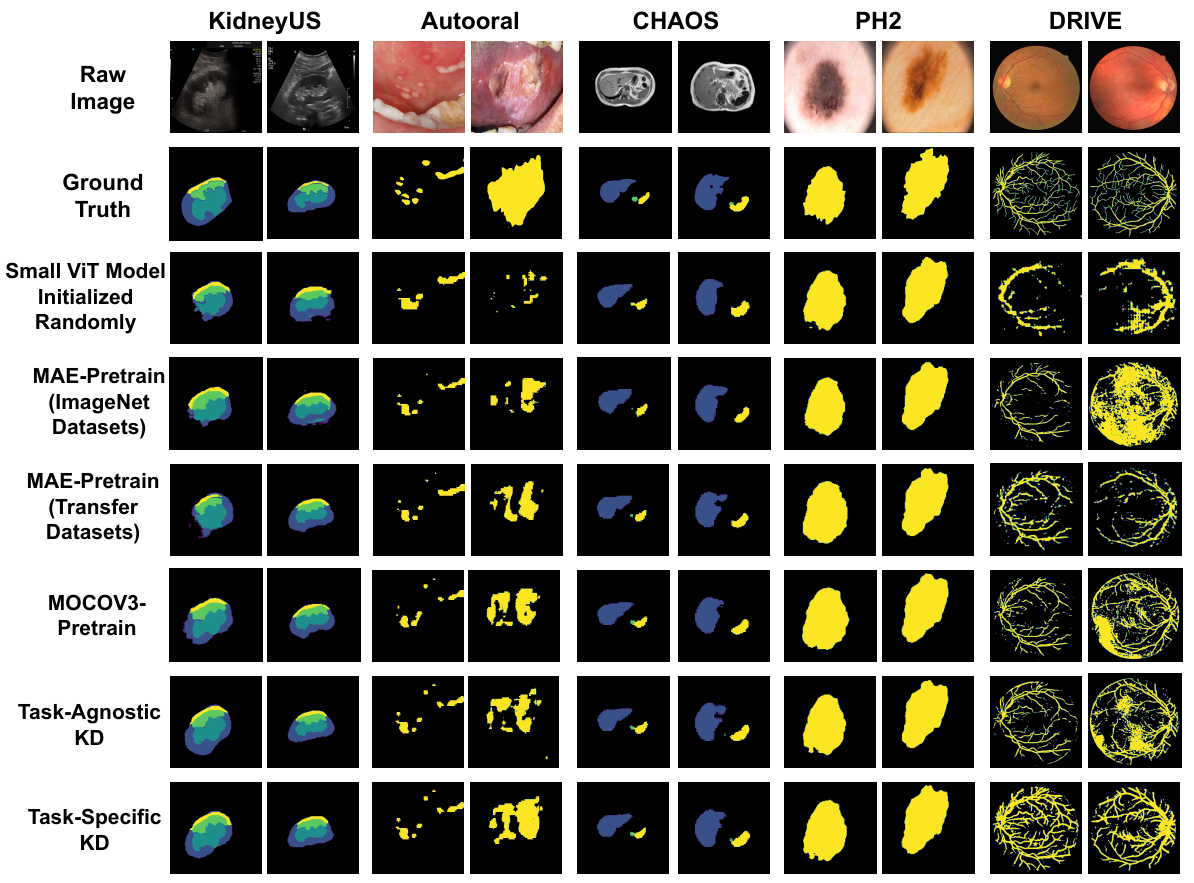}
    \caption{Qualitative comparison of segmentation results for a small Vision Transformer (ViT) model fine-tuned using different pretraining strategies across five medical image datasets (KidneyUS, Autooral, CHAOS, PH2, and DRIVE). Rows represent various pretraining and knowledge distillation strategies: Random Initialization, MAE-Pretrain (ImageNet Datasets), MAE-Pretrain (Transfer Datasets), MOCOV3-Pretrain (Transfer Datasets), Task-Agnostic KD (Transfer Datasets), and Task-Specific KD (Transfer Datasets). Each column corresponds to a dataset, with segmentation results visualized for representative samples. Task-Specific KD consistently produces segmentations that are more accurate and closely aligned with ground truth, demonstrating its effectiveness in leveraging task-specific features in medical image segmentation.}
    \label{fig9}
\end{figure*}

\begin{figure}[htbp]
    \centering
    \includegraphics[scale=0.42]{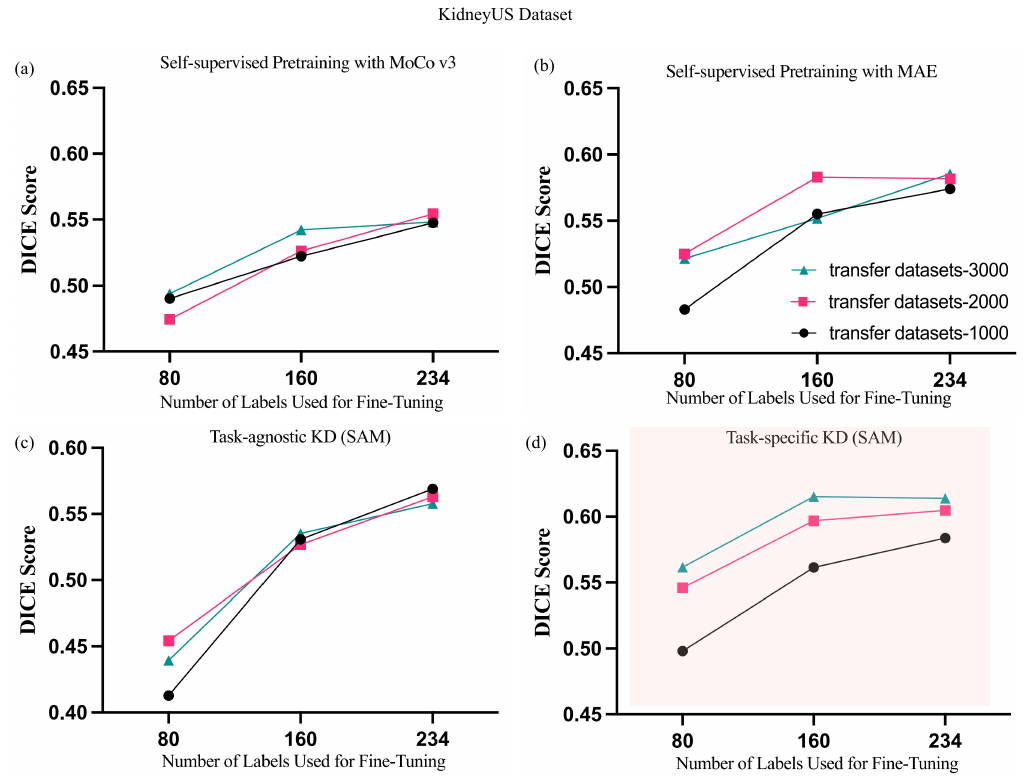}
    \caption{Segmentation performance on the KidneyUS dataset across different pre-training and knowledge distillation methods. (a) Dice score trends for self-supervised pretraining with MoCo v3 using transfer datasets of varying sizes (1000, 2000, and 3000 images). (b) Dice score trends for self-supervised pretraining with MAE using different transfer dataset sizes. (c) Dice score trends for task-agnostic KD (SAM) across different transfer dataset sizes and numbers of labeled samples used for fine-tuning. (d) Dice score trends for task-specific KD (SAM) show superior performance across all labeled sample sizes and transfer dataset scales, demonstrating the most significant performance improvements when larger transfer datasets are used.}
    \label{fig3}
\end{figure}

\begin{figure}[htbp]
    \centering
    \includegraphics[scale=0.42]{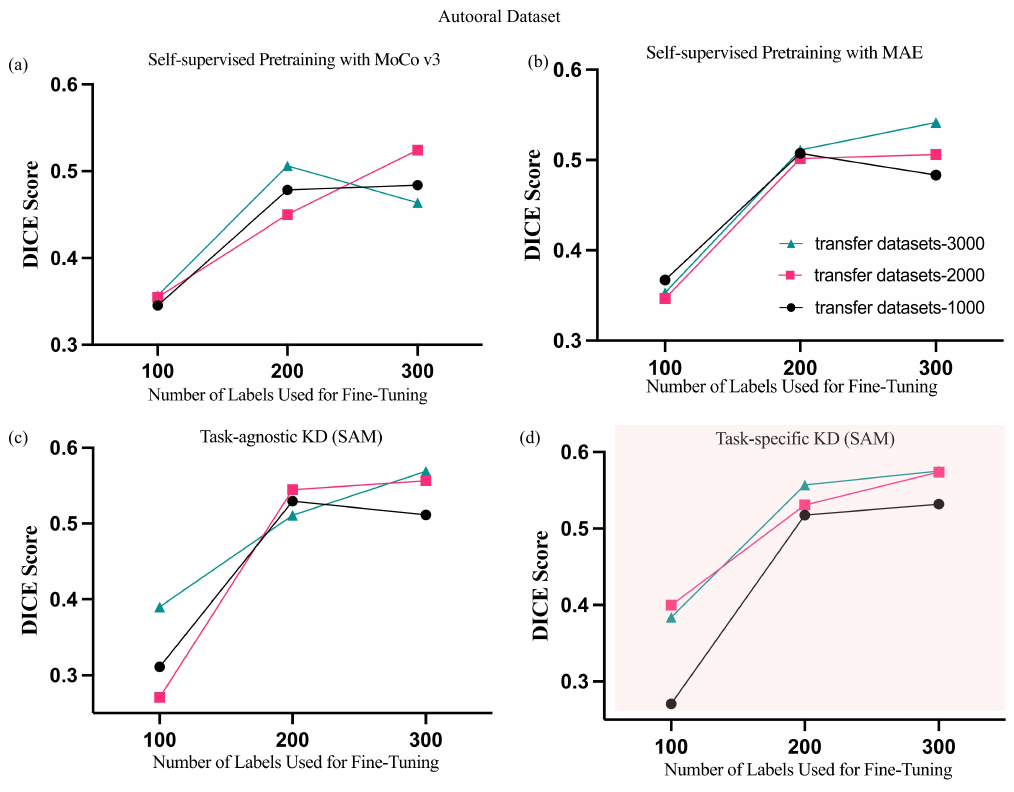}
    \caption{Segmentation performance on the Autooral dataset using various pre-training and knowledge distillation methods. (a) Dice score comparison across different numbers of labeled samples (100, 200, 300) for self-supervised pretraining with MoCo v3 using transfer datasets of varying sizes (1000, 2000, and 3000 images). (b-d) Dice score trends for each method under varying transfer dataset sizes, with Task-Specific KD consistently showing superior performance, especially with 300 labeled examples.}
    \label{fig4}
\end{figure}

\begin{figure}[htbp]
    \centering
    \includegraphics[scale=0.42]{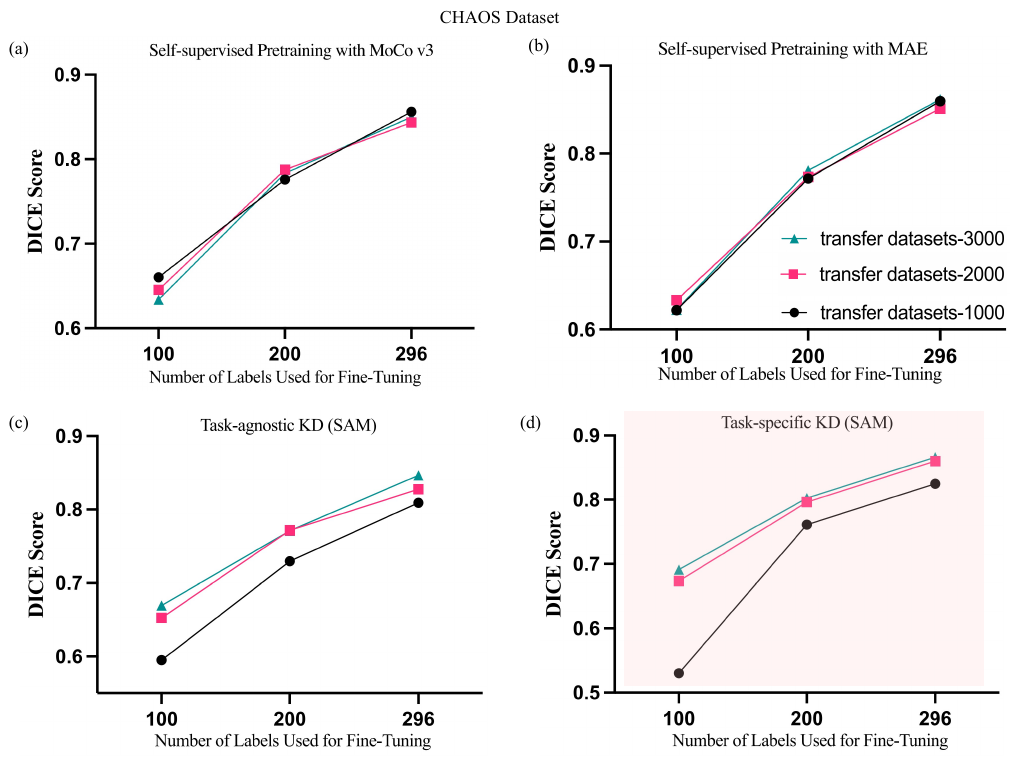}
    \caption{Segmentation performance on the CHAOS dataset for various pre-training and knowledge distillation methods. (a) Dice score comparison across different numbers of labeled samples (100, 200, 296) for self-supervised pretraining with MoCo v3 using transfer datasets of varying sizes (1000, 2000, and 3000 images). (b-d) Detailed Dice score trends for MAE, task-agnostic KD, and task-specific KD across different transfer dataset scales, illustrating that Task-Specific KD consistently delivers superior performance, especially with increasing labeled samples and larger transfer datasets.}
    \label{fig5}
\end{figure}

\begin{figure}[htbp]
    \centering
    \includegraphics[scale=0.52]{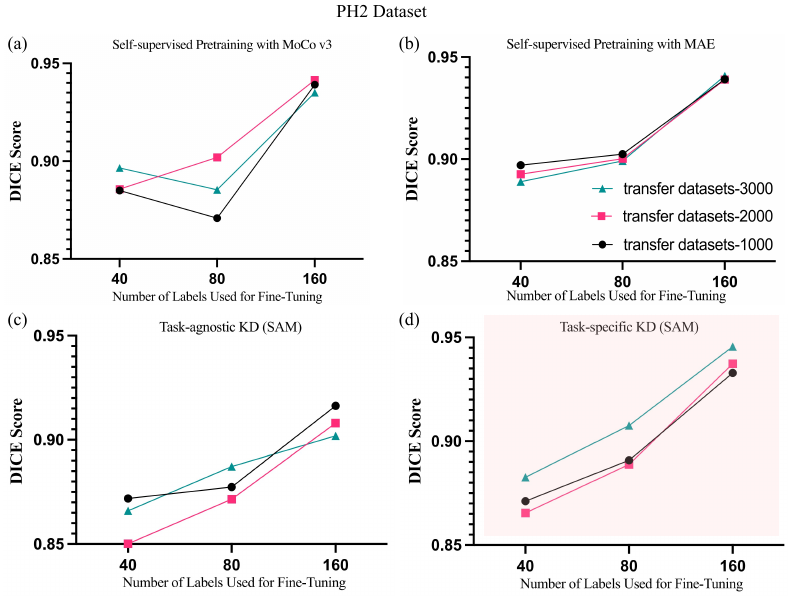}
    \caption{Performance comparison of different methods on the PH2 dataset for skin lesion segmentation, measured in Dice score. (a-d) Detailed analysis of self-supervised approaches (MoCo v3 and MAE) and knowledge distillation (task-agnostic and task-specific KD) across varying transfer dataset scales.}
    \label{fig6}
\end{figure}

\begin{figure}[htbp]
    \centering
    \includegraphics[scale=0.46]{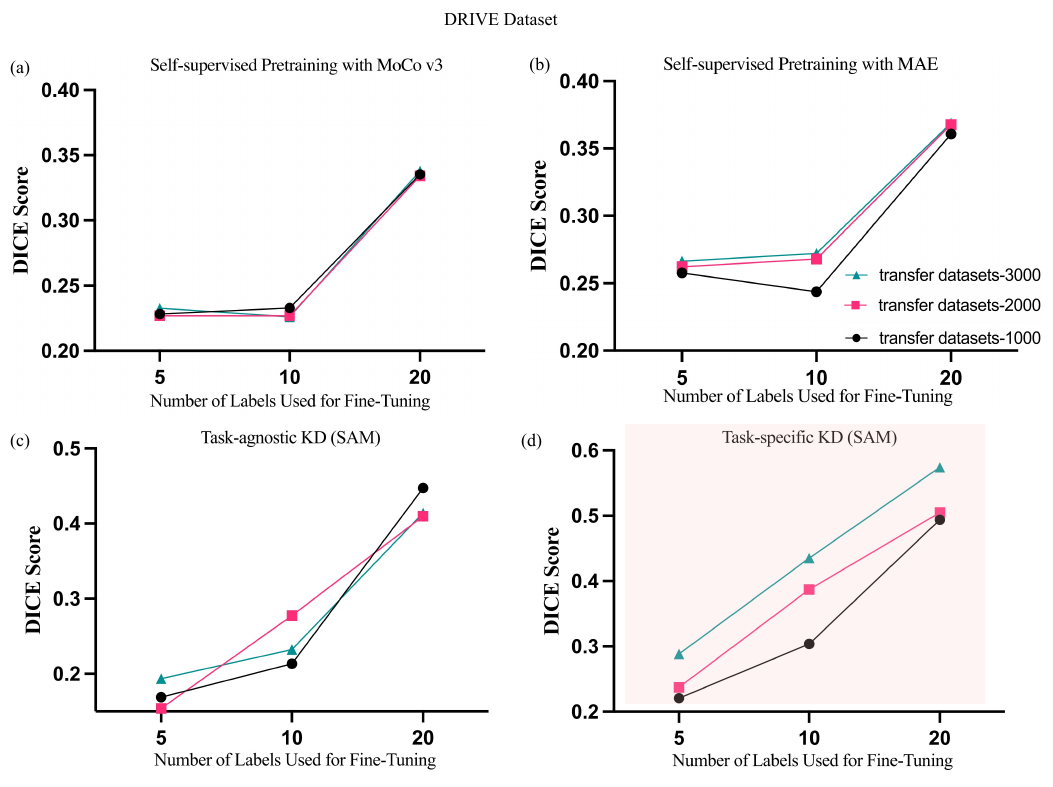}
    \caption{Performance comparison on the DRIVE dataset for retinal vessel segmentation, highlighting the effectiveness of various pre-training and knowledge distillation methods. (a-d) Trends in Dice score for self-supervised pre-training (MoCo v3, MAE) and knowledge distillation methods, show that Task-Specific KD provides the most substantial performance improvement, especially with limited labeled data. Task-specific KD also demonstrated superior boundary delineation as reflected by its lowest HD95 value.}
    \label{fig7}
\end{figure}

The effectiveness of our proposed Task-Specific Knowledge Distillation (KD) approach is thoroughly evaluated through extensive experiments on multiple medical image segmentation datasets, demonstrating notable performance improvements across all evaluated metrics and dataset types. Our method consistently outperformed both self-supervised and task-agnostic strategies, particularly under limited labeled data conditions, which is a common challenge in medical imaging. The results, summarized in Figure \ref{fig10}, Figure \ref{fig9}, Figures \ref{fig3} to \ref{fig7}, and Tables~\ref{taba1}--\ref{taba5}, demonstrate that transferring domain-specific knowledge from large pre-trained Vision Foundation Models (VFMs) significantly enhances segmentation quality, even when only a small number of labeled examples are available for fine-tuning.

As shown in Figure \ref{fig10}, the Dice scores for five datasets (KidneyUS, Autooral, CHAOS, PH2, and DRIVE) illustrate the consistent superiority of Task-Specific KD across different datasets and label counts. Each subplot in Figure \ref{fig10} (a)-(e) visualizes segmentation performance trends, with Task-Specific KD consistently achieving the highest Dice scores compared to other methods, including task-agnostic KD and self-supervised approaches (MoCo v3, MAE). This advantage is particularly evident as the number of labeled samples increases, suggesting that Task-Specific KD effectively adapts the pre-trained model to target-specific features, enabling robust performance gains even under limited supervision.

Figures \ref{fig3} to \ref{fig7} further illustrate the robustness and adaptability of Task-Specific KD in various experimental settings, including different scales of transfer datasets and varying numbers of labeled samples for fine-tuning. Unlike other pre-training or KD methods, Task-Specific KD shows a stable improvement trend in segmentation performance as both the transfer dataset size and the number of fine-tuning labels increase. Across most datasets, TSKD consistently achieves higher Dice scores, with clear advantages observed as the transfer dataset size expands from 1000 to 3000 synthetic images, and as the labeled sample count rises from minimal to full scale.

The qualitative results in Figure \ref{fig9} further illustrate the impact of different pretraining strategies on the segmentation performance of a small Vision Transformer (ViT) model across diverse medical image datasets. Task-specific KD captures fine-grained anatomical details more accurately than other methods, which is essential for achieving high performance in medical segmentation tasks.

\textbf{KidneyUS Dataset.} The experimental results on the KidneyUS dataset (Table \ref{taba1} and Figure \ref{fig3}) demonstrate the superiority of Task-Specific KD over other methods across varying transfer dataset sizes and labeled samples. Task-Specific KD achieved the highest mean Dice score of 0.6139 when using 3000 transfer images and 234 labeled examples, outperforming task-agnostic KD, MoCo v3, and MAE-based pretraining methods. For all segmentation regions—Capsule, Medulla, Central Echo Complex, and Cortex—Task-Specific KD consistently provided better Dice scores and HD95 metrics. For instance, in the Medulla, Task-Specific KD achieved a Dice score of 0.7615, notably better than the corresponding scores of task-agnostic KD (Dice: 0.7104) and MAE (Dice: 0.7275). With fewer labeled examples (80 or 160), Task-Specific KD maintained a performance edge. For example, with 160 labeled examples and 1000 transfer images, it achieved a mean Dice score of 0.5615, compared to 0.5307 for task-agnostic KD and 0.5550 for MAE, emphasizing the robustness of Task-Specific KD across different data scales. Additionally, Task-Specific KD scaled more effectively with increasing transfer dataset sizes. Dice scores consistently improved with larger transfer datasets, rising from 0.4980 with 1000 images to 0.5616 with 3000 images (using 80 labeled samples during the fine-tuning phase), a trend less evident in other methods, which saw diminishing gains.

\textbf{Autooral Dataset.} The Task-Specific KD approach achieved the highest segmentation performance on the Autooral dataset, as depicted in Figure \ref{fig4} and Table~\ref{taba2}. With a Dice score of 0.5754 using 300 labeled examples and 3000 transfer images, outperforming all other methods, including Task-Agnostic KD (Dice: 0.5688), MoCo v3 (Dice: 0.4636), and MAE (Dice: 0.5416). This demonstrates the advantage of task-specific KD in capturing domain-relevant features, particularly for ulcer segmentation.
With 200 labeled examples, Task-Specific KD also showed superior results, scoring 0.5570, compared to 0.5109 for Task-Agnostic KD and 0.5112 for MAE. Even with 100 labeled examples, it maintained a competitive Dice score of 0.3840, slightly outperforming MoCo v3 and MAE, showcasing robustness in data-limited settings.
The performance of Task-Specific KD improved with larger transfer datasets, increasing from 0.5320 with 1000 transfer images to 0.5754 with 3000 transfer images (using 300 labeled samples during the fine-tuning phase), while other methods showed diminishing returns. Additionally, it demonstrated better boundary precision, with an HD95 of 9.57 mm, compared to Task-Agnostic KD (13.22 mm), further highlighting its ability to capture fine anatomical details.

\textbf{CHAOS Dataset.} Task-Specific KD achieved the highest segmentation performance on the CHAOS dataset, with a Dice score of 0.8663 using 296 labeled samples, surpassing Task-Agnostic KD (0.8465) and other self-supervised methods (Table \ref{taba3}, Figure \ref{fig5}). Task-Specific KD excelled in segmenting complex organs, achieving superior Dice scores for the liver (0.8955), left kidney (0.8680), and spleen (0.8463), outperforming Task-Agnostic KD across these structures. 
Scalability was another strength: Task-specific KD improved with larger transfer datasets, with Dice scores rising from 0.8248 (1000 images) to 0.8663 (3000 images) when using 296 labeled samples. It also demonstrated better boundary precision, with a mean HD95 of 4.19 mm, outperforming Task-Agnostic KD (4.53 mm). Even with fewer labeled samples (100 or 200), Task-Specific KD maintained its performance edge, delivering robust results in challenging segmentation tasks. 

\textbf{PH2 Dataset.} For skin lesion segmentation on the PH2 dataset, Task-Specific KD demonstrated the most substantial performance boost, achieving a Dice score of 0.9455 with 160 labeled samples and a transfer dataset size of 3000, as seen in Figure \ref{fig6} and Table~\ref{taba4}. This score significantly exceeded the performance of models trained from scratch or using task-agnostic KD, indicating that leveraging domain-specific knowledge distillation is particularly beneficial in dermatological tasks where fine-grained lesion details are crucial for accurate diagnosis.

\textbf{DRIVE Dataset.} Our experiments on the DRIVE dataset, aimed at retinal vessel segmentation, showed that Task-Specific KD reached a Dice score of 0.5741, outperforming other pre-training methods, including self-supervised and task-agnostic KD, by a substantial margin. The model achieved the lowest HD95 value of 3.8028 mm, indicative of superior boundary delineation performance, as illustrated in Figure \ref{fig7} and Table~\ref{taba5}. The ability to maintain high accuracy in vessel segmentation, even with limited labeled data, underscores the capability of our Task-Specific KD framework to effectively transfer detailed semantic information, which is crucial for vascular structure analysis.

\textbf{Comparison Across Datasets.} Across all datasets, Task-Specific KD demonstrated superior performance, particularly in low-data regimes where data scarcity is most problematic. The consistent improvements in Dice, HD95, and mean Intersection over Union (mIoU) metrics emphasize the importance of tailoring pre-trained VFM to specific target tasks. The domain-specific adaptation facilitated by LoRA fine-tuning played a crucial role in enhancing the segmentation performance of our student model, thereby supporting efficient deployment without compromising accuracy.

The experimental results underscore the efficacy of Task-Specific KD combined with LoRA fine-tuning in significantly boosting segmentation performance across diverse medical imaging tasks. By effectively leveraging the domain-specific knowledge captured by VFM, our approach provides a robust solution for specific medical image segmentation, demonstrating its potential for broader application in clinical practice.

\subsubsection{Importance of Transfer Set Size}

The size of the transfer dataset plays a crucial role in the performance of pre-training strategies for medical image segmentation, with our experiments highlighting three key insights. 
First, larger transfer sets resulted in improved segmentation performance across different methods and datasets.
For instance, in the DRIVE dataset, Task-Specific KD with 3000 transfer images and 20 fine-tuning labels achieved a 16.22\% improvement in the Dice score compared to using 1000 transfer images.
Similarly, in the KidneyUS dataset, Task-Specific KD achieved a Dice score of 0.6139 using 3000 transfer images and 234 fine-tuning labels, compared to 0.5837 with 1000 transfer images. These results underscore the importance of larger transfer sets in learning robust feature representations, especially for complex segmentation tasks such as retinal vessel and kidney images.

However, we also observed diminishing returns beyond a certain dataset size. For example, in the Autooral dataset, the Dice score improvement from 2000 to 3000 transfer images was significantly smaller than the leap from 1000 to 2000 images. This suggests that while expanding transfer dataset size can enhance performance, the gains eventually plateau as the model reaches its optimal learning capacity, indicating the need for a practical balance between dataset size and computational cost.
Interestingly, knowledge distillation methods, particularly Task-Specific KD, demonstrated notable efficiency even with smaller transfer sets. In the PH2 dataset, Task-Specific KD achieved a Dice score of 0.9329 with just 1000 transfer images (160 fine-tuning labels), which was close to the performance of 0.9455 achieved with 3000 images. This indicates that leveraging knowledge distillation from a fine-tuned large model, such as SAM, can effectively bridge the gap in data-scarce scenarios, making it a powerful strategy for medical applications where labeled data is often limited.

\subsubsection{Training Efficiency}

\begin{table*}[!ht]
\setlength{\abovecaptionskip}{0pt}
\setlength{\belowcaptionskip}{0pt}
\centering
% \small
\normalsize
\caption{ Comparison of pre-training methods and GPU hours on segmentation performance across multiple medical datasets.}
\scalebox{0.65}{
\begin{tabular}{ccc
S[round-precision=4,round-mode=places]
S[round-precision=4,round-mode=places]}
\hline
\textbf{Target Task}      & \textbf{Method}                      & \textbf{GPU Hours (Pre-training)} & \textbf{Performance (Mean Dice)} & \textbf{Performance Improvement (\%)} \\ \hline
\multirow{5}{*}{kidneyus} & MAE-Pretrain (ImageNet Datasets)     & 23h*                              & 0.544895                         & 0.825448                              \\
                          & MAE-Pretrain (Transfer Datasets)     & 2.5h                              & 0.585460                          & 8.331452                              \\
                          & MOCOV3-Pretrain (Transfer Datasets)  & 30.6h                             & 0.548090                          & 1.416639                              \\
                          & Task-Agnostic KD (Transfer Datasets) & 16h                               & 0.557675                         & 3.190214                              \\
                          \rowcolor{lightblue}
                          & Task-Specific KD (Transfer Datasets) & 16.2h                             & 0.613858                         & 13.586118                    \\ \hline
\multirow{5}{*}{Autooral} & MAE-Pretrain (ImageNet Datasets)     & 23h*                              & 0.549543                         & 22.936418                             \\
                          & MAE-Pretrain (Transfer Datasets)     & 0.9h                              & 0.541607                         & 21.161082                             \\
                          & MOCOV3-Pretrain (Transfer Datasets)  & 3.6h                              & 0.463555                         & 3.700332                              \\
                          & Task-Agnostic KD (Transfer Datasets) & 2.8h                              & 0.568807                         & 27.245903                             \\
                          \rowcolor{lightblue}
                          & Task-Specific KD (Transfer Datasets) & 3h                                & 0.575414                         & 28.723933                    \\ \hline
\multirow{5}{*}{Chaos}    & MAE-Pretrain (ImageNet Datasets)     & 23h*                              & 0.858229                         & 5.77399                               \\
                          & MAE-Pretrain (Transfer Datasets)     & 0.9h                              & 0.861574                         & 6.186251                              \\
                          & MOCOV3-Pretrain (Transfer Datasets)  & 3.6h                              & 0.849804                         & 4.735636                              \\
                          & Task-Agnostic KD (Transfer Datasets) & 2.8h                              & 0.846488                         & 4.326949                              \\
                          \rowcolor{lightblue}
                          & Task-Specific KD (Transfer Datasets) & 3h                                & 0.866252                         & 6.762799                     \\ \hline
\multirow{5}{*}{PH2}      & MAE-Pretrain (ImageNet Datasets)     & 23h*                              & 0.926610                          & 0.986091                              \\
                          & MAE-Pretrain (Transfer Datasets)     & 2.5h                              & 0.940803                         & 2.532908                              \\
                          & MOCOV3-Pretrain (Transfer Datasets)  & 30.6h                             & 0.935063                         & 1.907337                              \\
                          & Task-Agnostic KD (Transfer Datasets) & 16h                               & 0.901878                         & -1.709312                             \\
                          \rowcolor{lightblue}
                          & Task-Specific KD (Transfer Datasets) & 16.2h                             & 0.945516                         & 3.046552                     \\ \hline
\multirow{5}{*}{DRIVE}    & MAE-Pretrain (ImageNet Datasets)     & 23h*                              & 0.390915                         & 19.039499                             \\
                          & MAE-Pretrain (Transfer Datasets)     & 2.5h                              & 0.368881                         & 12.329814                             \\
                          & MOCOV3-Pretrain (Transfer Datasets)  & 30.6h                             & 0.337846                         & 2.879190                               \\
                          & Task-Agnostic KD (Transfer Datasets) & 16h                               & 0.414178                         & 26.123432                             \\
                          \rowcolor{lightblue}
                          & Task-Specific KD (Transfer Datasets) & 16.2h                             & 0.574099                         & 74.821783                    \\ \hline

\label{tab1}
\end{tabular}}
\vspace{-.1in}
\end{table*}

In this section, we compare the training efficiency and segmentation performance across five medical imaging datasets using various pre-training strategies. The methods evaluated include MAE pre-trained on ImageNet, MAE pre-trained on transfer datasets, MoCo v3 pre-training, Task-Agnostic Knowledge Distillation (KD), and Task-Specific KD. The experimental results, along with GPU training times, are detailed in Table~\ref{tab1}.

\paragraph{\textbf{Pre-training on ImageNet (MAE).}}
For the MAE pre-training on ImageNet, we utilized pre-trained weights and training times as reported~\cite{wang2023closer}. Specifically, the MAE model was trained for 400 epochs on ImageNet using a system equipped with 8 NVIDIA V100 GPUs, which required approximately 23 hours. Subsequently, the model was fine-tuned for each medical imaging task using the maximum number of labeled samples available. Despite yielding competitive performance, this approach required substantial computational resources due to the extensive ImageNet pre-training.

\paragraph{\textbf{Pre-training on Transfer Datasets.}}
For the remaining methods—MAE pre-trained on transfer datasets, MoCo v3 pre-training, Task-Agnostic KD, and Task-Specific KD—we employed synthetic transfer datasets generated using a Swin-transformer-based diffusion model. These pre-training processes were executed on two NVIDIA L40 GPUs for 1600 epochs across all methods. The pre-training times varied based on the complexity of the strategy, with Task-Specific KD requiring slightly longer due to the additional task-specific knowledge distillation step.

\paragraph{\textbf{Comparison of Pre-training Methods.}}
The Task-Specific KD approach consistently achieved the best segmentation performance across all datasets, demonstrating substantial improvements, particularly in challenging tasks such as retinal vessel segmentation (DRIVE dataset), where it delivered a 74.82\% improvement. This significant gain can be attributed to the fine-tuning of the SAM model on each specific medical imaging task before knowledge distillation, thereby optimizing both the encoder and decoder for the target task. The task-specific fine-tuning allowed the student model to benefit from more specialized feature representations and segmentation outputs, leading to superior performance.

Conversely, the Task-Agnostic KD method, which aligns only the encoder representations, demonstrated comparatively lower performance. This method relied on the original, unmodified SAM encoder without task-specific fine-tuning, resulting in less effective transfer of knowledge to the student model. In tasks requiring detailed and fine-grained segmentation, such as retinal vessel extraction, the relevance of the knowledge transferred was notably reduced. These findings emphasize the importance of fine-tuning large models such as SAM on the target domain before distillation, as this significantly enhances the relevance and utility of the transferred knowledge.

\paragraph{\textbf{GPU Training Time Analysis.}}
The MAE pre-training on ImageNet was the most computationally demanding, with a pre-training time of 23 hours. In contrast, methods utilizing transfer datasets required significantly less time, with the shortest pre-training time recorded at 0.9 hours for the Autooral dataset. Task-Specific KD required 16.2 hours of pre-training, making it more computationally intensive compared to other transfer dataset methods, but still considerably more efficient than ImageNet-based pre-training.
MoCo v3 pre-training, while competitive in certain tasks, exhibited the longest pre-training time among the transfer dataset methods at 30.6 hours, likely due to the larger batch size and the complexity of the contrastive learning objective.

\paragraph{\textbf{Performance and Computational Trade-offs.}}
Despite the computational overhead of Task-Specific KD, the associated performance gains justified the additional pre-training time. Compared to the MAE pre-training on ImageNet, Task-Specific KD achieved superior performance on most datasets while requiring significantly less pre-training time. This efficiency, combined with its ability to transfer task-specific knowledge effectively, positions it as a promising alternative to traditional large-scale pre-training approaches.
As summarized in Table \ref{tab1}, Task-Specific KD consistently delivered the best performance improvements, particularly in challenging segmentation tasks such as retinal vessel segmentation (DRIVE dataset) and kidney ultrasound (KidneyUS dataset), where capturing intricate details is critical. These results underscore the potential of Task-Specific KD to achieve state-of-the-art segmentation performance while reducing computational demands.

\subsubsection{Memory Efficiency and Performance of ViT-Tiny}

One of the key advantages of Task-Specific Knowledge Distillation (KD) is the ability to compress large models, like SAM, into smaller models, such as ViT-Tiny, without significant loss in segmentation performance. Table \ref{tab:sam_finetuning} compares the memory footprint and segmentation accuracy (Dice scores) of the SAM-LoRA model, fine-tuned with the smallest labeled dataset, and the ViT-Tiny model was pre-trained using Task-Specific Knowledge Distillation (TSKD) on a transfer dataset consisting of 3000 synthetic images, followed by fine-tuning with the minimum number of labeled samples available for each dataset.

\begin{table*}[htbp]
\centering
% \small
% \normalsize
\large
\caption{Comparison of memory and segmentation performance across different datasets for SAM-LoRA (fine-tuned on the smallest labeled dataset) and ViT-Tiny (trained with Task-Specific KD). ViT-Tiny was trained using a transfer dataset of 3000 synthetic images and fine-tuned with the smallest labeled dataset.}
\scalebox{0.46}{
\label{tab:sam_finetuning}
\begin{tabular}{cccccc}
\hline
\textbf{Dataset}  & \textbf{SAM Size (MB)} & \textbf{LoRA SAM Extra Memory (MB)} & \textbf{ViT-Tiny + Seg Head (MB)} & \textbf{SAM-LoRA Dice (Small Label Set)} & \textbf{ViT-Tiny Dice (Task-Specific KD)} \\
\hline
KidneyUS & 357 & 17.2 & 33.5 & 0.5499 & 0.5616 \\
Autooral & 357 & 15.3 & 33.0 & 0.7056 & 0.3840 \\
CHAOS    & 357 & 18.8 & 33.0 & 0.6799 & 0.6912 \\
PH2      & 357 & 16.1 & 33.5 & 0.8786 & 0.8826 \\
DRIVE    & 357 & 16.1 & 33.5 & 0.7083 & 0.2886 \\
\hline
\end{tabular}}
\end{table*}

The results in Table \ref{tab:sam_finetuning} demonstrate that ViT-Tiny, trained using Task-Specific KD, offers significant memory savings while achieving competitive segmentation performance. With only 33 MB of memory, ViT-Tiny achieves Dice scores close to, and sometimes even surpasses, the much larger SAM-LoRA model. For example, in the CHAOS dataset, ViT-Tiny slightly outperformed SAM-LoRA, achieving a Dice score of 0.6912 compared to SAM-LoRA’s 0.6799, while consuming nearly ten times less memory.

In the PH2 dataset, which involves skin lesion segmentation, ViT-Tiny achieved a Dice score of 0.8826, almost identical to SAM-LoRA’s 0.8786, further highlighting the effectiveness of KD in transferring task-specific knowledge from a large teacher model to a small student model without sacrificing segmentation accuracy.

For more complex datasets such as Autooral and DRIVE, where anatomical details or fine structures are more challenging to segment, SAM-LoRA's higher capacity led to noticeably better performance. For instance, in the Autooral dataset, SAM-LoRA achieved a Dice score of 0.7056, significantly higher than ViT-Tiny’s 0.3840. This suggests that while ViT-Tiny is highly memory-efficient, it may struggle with more intricate segmentation tasks when compared to its larger counterpart.

Despite these trade-offs, Task-Specific KD enables ViT-Tiny to achieve competitive performance with minimal memory usage, making it a strong candidate for resource-limited applications in medical image analysis.

\subsection{Ablation Study}

\subsubsection{Impact of Task-Specific KD Loss Weights}

\begin{table*}[htbp]
\centering
% \small
% \normalsize
\large
\caption{
Summary of KD loss functions, mask processing, and loss weights in task-specific KD experiments. Each experiment typically includes two types of KD: (1) KD applied at the decoder level, focusing on aligning segmentation outputs (e.g., using MSE or CrossEntropy for spatial alignment of masks); and (2) KD applied at the encoder level, where hidden states from the teacher and student models are aligned to transfer rich feature representations. However, in TS-KD6, only decoder-level KD (MSE) is applied, with no hidden states distillation. The table also details how SAM's low-resolution logits were processed (interpolated or uninterpolated) and specifies the loss weights assigned to each KD term, balancing the contributions of spatial alignment and feature representation transfer.
}
\scalebox{0.37}{
\begin{tabular}{l l l l l l}
\toprule
\textbf{Experiment Group} & \textbf{Loss Type}        & \textbf{Mask Processing}                                 & \textbf{Hidden States Distillation} & \textbf{Loss Weights}                                 & \textbf{Features}                                 \\ 
\midrule
\textbf{TS-KD1}          & MSE + Hidden Loss         & Removing the last dimension of masks                     & Yes                                & MSE weight: 0.2                                      & Emphasizes Hidden States alignment; MSE for mask alignment  \\
                          &                          &                                                           &                                    & Hidden Loss weight: 1.0                               &                                                \\ 
\textbf{TS-KD2}          & MSE + Hidden Loss         & Low-res logits interpolated to student output size         & Yes                                & MSE weight: 0.1                                      & Aligns low-res logits, reduces high-res dependency          \\ 
                          &                          &                                                           &                                    & Hidden Loss weight: 1.0                               &                                                \\ 
\textbf{TS-KD3}          & MSE + Hidden Loss         & Uninterpolated low-res logits                             & Yes                                & MSE weight: 0.001                                    & Removes interpolation, low-weight MSE loss                   \\ 
                          &                          &                                                           &                                    & Hidden Loss weight: 1.0                               &                                                \\ 
\textbf{TS-KD4}          & CrossEntropy + Hidden Loss & CrossEntropy loss aligns class probability distribution   & Yes                                & CrossEntropy weight: 1.0                             & Focuses on class alignment, using CrossEntropy instead of MSE  \\ 
                          &                          &                                                           &                                    & Hidden Loss weight: 1.0                               &                                                \\ 
\textbf{TS-KD5}          & CrossEntropy + Hidden Loss & CrossEntropy loss aligns class probability distribution   & Yes, reduced weight                & CrossEntropy weight: 1.0                             & Similar to TS-KD4 but with reduced Hidden States weight      \\ 
                          &                          &                                                           &                                    & Hidden Loss weight: 0.1                               &                                                \\ 
\textbf{TS-KD6}          & MSE                       & Interpolated low-res logits                               & No                                 & MSE weight: 0.1                                      & No Hidden Loss, focuses on MSE loss                           \\ 
                          &                          &                                                           &                                    & Hidden Loss weight: 0.0                               &                                                \\ 
\textbf{TS-KD7}          & MSE + Hidden Loss         & Interpolated low-res logits                               & Yes                                & MSE weight: 0.1                                      & Restores interpolation, moderate MSE alignment                \\ 
                          &                          &                                                           &                                    & Hidden Loss weight: 0.1                               &                                                \\ 
                          \rowcolor{lightblue}
\textbf{TS-KD8}          & MSE + Hidden Loss         & Interpolated low-res logits                               & Yes                                & MSE weight: 0.2                                      & Similar to TS-KD7 but with increased MSE loss weight         \\ \rowcolor{lightblue}
                          &                          &                                                           &                                    & Hidden Loss weight: 0.1                               &                                                \\ 
\bottomrule
\end{tabular}}
\label{tab:kd_experiments}
\end{table*}

\begin{table*}[!ht]
\setlength{\abovecaptionskip}{0pt}
\setlength{\belowcaptionskip}{0pt}
\centering
% \small
% \normalsize
\large
\caption{Segmentation performance (Dice score and HD95) across different organs in the CHAOS dataset for various task-specific KD configurations (TS-KD1 to TS-KD8). The experiments utilize a training dataset that includes 3000 synthetic images for pre-training and 296 labeled real images for fine-tuning.}
\scalebox{0.41}{
\begin{tabular}{ccc
S[round-precision=4,round-mode=places]
S[round-precision=4,round-mode=places]
S[round-precision=4,round-mode=places]
S[round-precision=4,round-mode=places]
S[round-precision=4,round-mode=places]
S[round-precision=4,round-mode=places]
S[round-precision=4,round-mode=places]
S[round-precision=4,round-mode=places]
S[round-precision=4,round-mode=places]
S[round-precision=4,round-mode=places]}
\hline
\multirow{2}{*}{\textbf{Method Description}} & \multirow{2}{*}{\textbf{Transfer Data Scale}} & \multirow{2}{*}{\textbf{Number of Labels}} & \multicolumn{2}{c}{\textbf{Liver}} & \multicolumn{2}{c}{\textbf{Right kidney}} & \multicolumn{2}{c}{\textbf{Left kidney}} & \multicolumn{2}{c}{\textbf{Spleen}} & \multicolumn{2}{c}{\textbf{Mean}}      \\  
                                             &                                               &                                            & \textbf{Dice} & \textbf{HD95 (mm)} & \textbf{Dice}     & \textbf{HD95 (mm)}    & \textbf{Dice}    & \textbf{HD95 (mm)}    & \textbf{Dice}  & \textbf{HD95 (mm)} & \textbf{Dice}     & \textbf{HD95 (mm)} \\ \hline
TS-KD1                                      & 3000                                          & 296                                        & 0.880780      & 2.512817           & 0.811078          & 15.695086             & 0.829195         & 3.278027              & 0.789511       & 2.988633           & 0.827641          & 6.118641           \\
TS-KD2                                      & 3000                                          & 296                                        & 0.894059      & 1.892648           & 0.846015          & 15.664497             & 0.845397         & 2.748970              & 0.797799       & 3.052793           & 0.845818          & 5.839727           \\
TS-KD3                                      & 3000                                          & 296                                        & 0.893587      & 2.096772           & 0.856283          & 11.796230             & 0.832380         & 2.641997              & 0.773800       & 2.976225           & 0.839013          & 4.877806           \\
TS-KD4                                      & 3000                                          & 296                                        & 0.895091      & 1.919379           & 0.840468          & 11.811757             & 0.831215         & 2.237619              & 0.792454       & 4.379009           & 0.839807          & 5.086941           \\
TS-KD5                                      & 3000                                          & 296                                        & 0.898380      & 1.824045           & 0.873824          & 16.755830             & 0.873780         & 1.499861              & 0.818503       & 2.036326           & 0.866122          & 5.529016           \\
TS-KD6                                      & 3000                                          & 296                                        & 0.893109      & 1.875470           & 0.827186          & 11.142326             & 0.843632         & 2.007660              & 0.820420       & 2.447642           & 0.846087          & 4.368274           \\
TS-KD7                                      & 3000                                          & 296                                        & 0.894521      & 1.802584           & 0.831130          & 11.392146             & 0.852334         & 2.643276              & 0.802051       & 2.701805           & 0.845009          & 4.634953           \\
\rowcolor{lightblue}
\textbf{TS-KD8}                             & 3000                                          & 296                                        & 0.895469      & 2.077859           & 0.855298          & 11.802779             & 0.867955         & 1.369500              & 0.846288       & 1.495408           & 0.866252 & 4.186386  \\ \hline
\label{tab4}
\end{tabular}}
\vspace{-.1in}
\end{table*}

The results of our experiments on task-specific knowledge distillation (KD) demonstrate the effectiveness of fine-tuning the SAM model and transferring its domain-specific knowledge to a smaller Vision Transformer (ViT) model for medical image segmentation tasks. By leveraging both MSE and Hidden States distillation losses, we achieved notable improvements in segmentation accuracy across various anatomical structures within the CHAOS dataset, as detailed in Table~\ref{tab:kd_experiments} and Table~\ref{tab4}.

In particular, the experiment group \textbf{TS-KD8}, which employed a higher MSE weight of 0.2 alongside a moderate Hidden Loss weight of 0.1, achieved the best overall performance, with a mean Dice score of \textbf{0.8663} and an HD95 of \textbf{4.19 mm}. This configuration balanced the contributions of spatial alignment and feature representation transfer, proving especially effective for segmenting complex anatomical structures like the liver and kidneys. For instance, in the case of the left kidney, \textbf{TS-KD8} achieved a Dice score of \textbf{0.8680} with an HD95 of \textbf{1.37 mm}, outperforming other configurations in terms of both segmentation accuracy and boundary precision.

Comparatively, the lower MSE weight in \textbf{TS-KD7} (MSE: 0.1, Hidden Loss: 0.1) resulted in a slightly reduced Dice score of $0.8450$, with an HD95 of  $4.63 mm$. This suggests that while moderate MSE weighting helps maintain reasonable segmentation quality, increasing the MSE weight yields more precise spatial alignment, especially in structures with intricate boundaries, such as the spleen, where TS-KD8 recorded a Dice score of $0.8463$ compared to $0.8021$ in TS-KD7.

The role of Hidden States distillation was further highlighted by the comparison with TS-KD6, which did not employ Hidden Loss and relied solely on MSE. This configuration, although achieving a competitive Dice score of $0.8461$, lagged in boundary precision with an HD95 of $4.37 mm$. The absence of Hidden Loss in TS-KD6 likely reduced the model’s ability to transfer rich feature representations from the teacher to the student, resulting in slightly inferior performance, especially in structures like the right kidney, where TS-KD6 recorded an HD95 of $11.14 mm$, compared to $11.80 mm$ in TS-KD8.

In experiments utilizing CrossEntropy loss (TS-KD4 and TS-KD5), the results indicate that CrossEntropy is less effective than MSE in this task-specific KD framework. Although TS-KD5, with reduced Hidden Loss weight (0.1), achieved a respectable Dice score of $0.8661$, it recorded an HD95 of $5.53 mm$, indicating that it struggled to maintain precise boundary delineation, particularly for the right kidney, which had an HD95 of $16.76 mm$. This finding supports the hypothesis that CrossEntropy, while effective in aligning class probabilities, is less suited to the fine-grained spatial alignment required for medical segmentation tasks, particularly in the CHAOS dataset.

The interpolation of SAM’s low-resolution logits to match the student model’s output size proved to be an essential technique for transferring useful knowledge from the teacher model. In TS-KD8, this interpolation, coupled with the MSE loss, allowed the student model to better capture the spatial characteristics of the target structures, leading to improved segmentation performance. For example, in the liver, TS-KD8 achieved a Dice score of $0.8955$ and an HD95 of $2.08 mm$, compared to $0.8808$ and $2.51 mm$, respectively, in TS-KD1, which used a simpler mask processing approach.

Furthermore, the experiments revealed diminishing returns when reducing the MSE weight excessively. TS-KD3, with an MSE weight of 0.001, produced a Dice score of $0.8390$ and an HD95 of $4.88 mm$, showing that an overly small MSE contribution undermines spatial alignment. This was evident in smaller structures like the spleen, where TS-KD3 recorded a Dice score of $0.7738$, compared to $0.8463$ in TS-KD8.

These results underscore the importance of balancing MSE and Hidden Loss in the KD framework. The configuration in \textbf{TS-KD8}, with its higher MSE weight, proved optimal for transferring spatial alignment and feature representations from the fine-tuned SAM model to the student ViT model. This combination of losses enabled the model to achieve superior segmentation accuracy and boundary precision, particularly in challenging structures within the CHAOS dataset. Task-specific KD, when fine-tuned with appropriately weighted losses, demonstrates substantial promise for improving medical image segmentation in data-limited scenarios.

\subsubsection{LoRA rank Impact of SAM Model Fine-tuning}

\begin{table*}[!ht]
\setlength{\abovecaptionskip}{0pt}
\setlength{\belowcaptionskip}{0pt}
\centering
% \small
% \normalsize
\large
\caption{Segmentation performance for selected LoRA ranks across different datasets and label counts.}
\scalebox{0.71}{
\begin{tabular}{ccc
S[round-precision=4,round-mode=places]
S[round-precision=4,round-mode=places]}
\hline
\textbf{Target Task}                              & \textbf{Lora rank} & \textbf{Number of Labels} & \textbf{Mean Dice} & \textbf{Mean HD95 (mm)} \\ \hline
\rowcolor{lightblue}
\multirow{3}{*}{SAM\&Lora Fine-Tuning (kidneyUS)} & 4                  & 80                        & 0.549917  & 12.725725      \\
                                                  & 4                  & 160                       & 0.601518           & 12.611959               \\
                                                  & 4                  & 234                       & 0.625375           & 13.291032               \\ \hline
                                                  \rowcolor{lightblue}
\multirow{3}{*}{SAM\&Lora Fine-Tuning (Autooral)} & 2                  & 100                       & 0.705630  & 19.917304      \\
                                                  & 2                  & 200                       & 0.794549           & 5.267413                \\
                                                  & 2                  & 300                       & 0.812124           & 4.186127                \\ \hline
                                                  \rowcolor{lightblue}
\multirow{3}{*}{SAM\&Lora Fine-Tuning (Chaos)}    & 16                 & 100                       & 0.679914  & 14.211168      \\
                                                  & 16                 & 200                       & 0.797582           & 5.406250                \\
                                                  & 16                 & 296                       & 0.865122           & 2.937228                \\ \hline
                                                  \rowcolor{lightblue}
\multirow{3}{*}{SAM\&Lora Fine-Tuning (PH2)}      & 8                  & 40                        & 0.878557  & 19.043960      \\
                                                  & 8                  & 80                        & 0.898359           & 17.434574               \\
                                                  & 8                  & 160                       & 0.967723           & 0.500000                \\ \hline
                                                  \rowcolor{lightblue}
\multirow{3}{*}{SAM\&Lora Fine-Tuning (DRIVE)}    & 8                  & 5                         & 0.708338  & 2.914214       \\
                                                  & 8                  & 10                        & 0.711920           & 3.000000                \\
                                                  & 8                  & 20                        & 0.759780           & 2.118034                \\ \hline

\label{tab2}
\end{tabular}}
\vspace{-.1in}
\end{table*}

\begin{figure*}[htbp]
    \centering
    \includegraphics[scale=0.61]{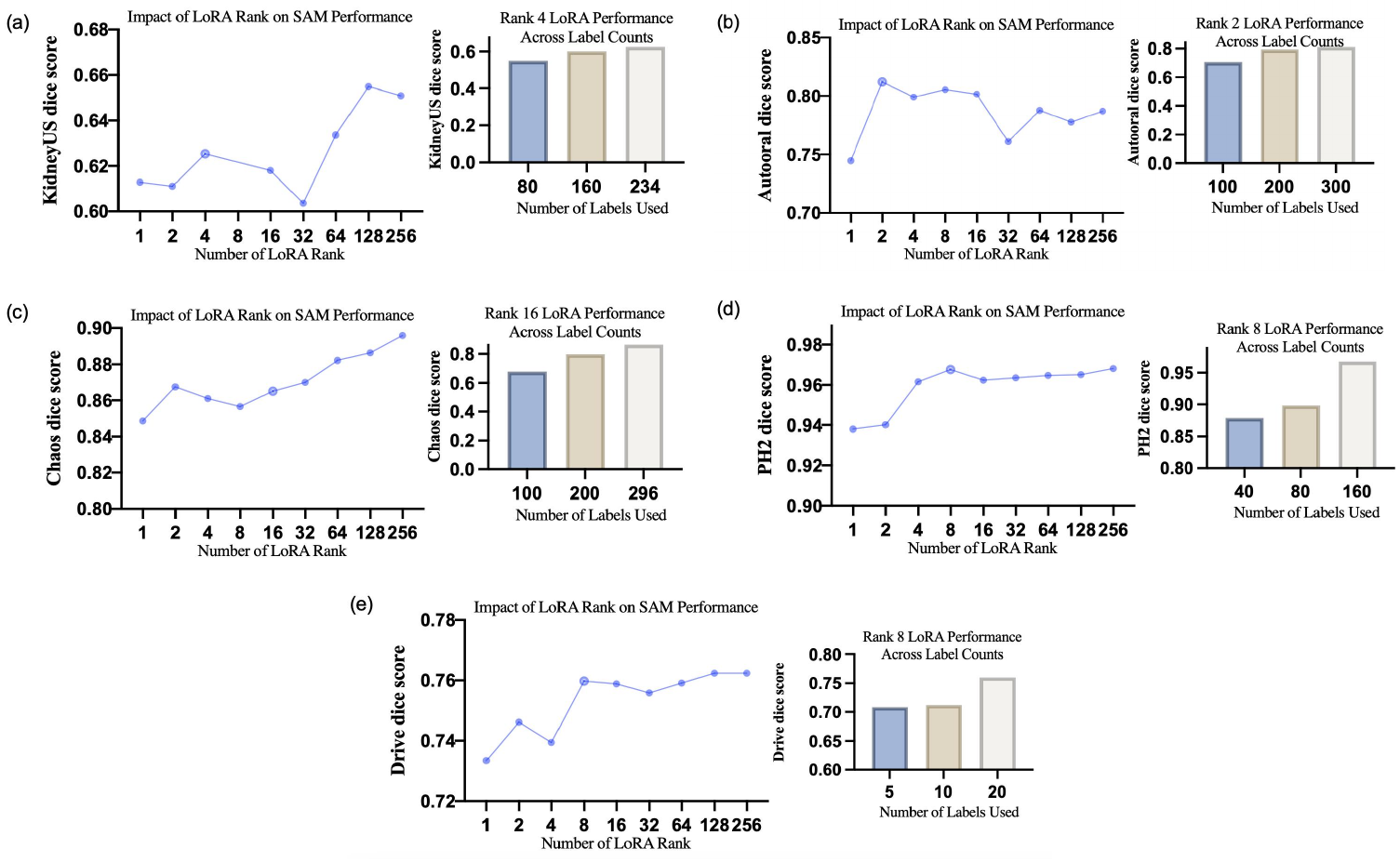}
    \caption{Influence of LoRA rank on SAM model segmentation performance across five datasets (KidneyUS, Autooral, CHAOS, PH2, and DRIVE). Each subfigure shows: (1) Left: Line plots displaying Dice scores as LoRA rank varies, illustrating how changes in rank affect segmentation accuracy per dataset. Datasets like CHAOS and PH2 benefit from higher ranks, while Autooral achieves optimal results with lower ranks. (2) Right: Bar charts showing Dice scores at the selected LoRA rank across different labeled sample sizes for each dataset. These results guided the selection of efficient LoRA ranks for task-specific SAM fine-tuning, later applied in knowledge distillation (KD) with SAM as the teacher model.}
    \label{fig8}
\end{figure*}

To investigate the influence of different LoRA ranks on the fine-tuning of the SAM model, we conducted experiments across a variety of datasets, employing a range of LoRA ranks. Our objective was to identify a LoRA rank that strikes an optimal balance between segmentation performance and model efficiency, which would later serve as the foundation for task-specific knowledge distillation (KD). Figure \ref{fig8} presents the segmentation performance of the SAM model at various LoRA ranks, evaluated using the Dice score on datasets such as KidneyUS, Autooral, CHAOS, PH2, and DRIVE.

\paragraph{\textbf{LoRA Rank Selection and Performance.}}

The experiments highlight that LoRA rank selection significantly affects the fine-tuning performance of the SAM model (as shown in Figure \ref{fig8} and Table \ref{tab2}). Detailed results can be found in Appendix Tables~\ref{taba6}--\ref{taba10}:

\begin{itemize}
    \item \textbf{KidneyUS Dataset (Figure \ref{fig8}a).} The results reveal a steady improvement in the Dice score with increasing LoRA rank, reaching a peak at rank 128. However, for subsequent experiments, a lower rank (Rank 4) was selected to achieve a competitive performance (0.6254 Dice score with 234 labels) while enhancing computational efficiency.
    
    \item \textbf{Autooral Dataset (Figure \ref{fig8}b).} A LoRA rank of 2 delivered strong segmentation performance, achieving a Dice score of 0.7056 with only 100 labeled images. This rank was subsequently chosen for KD experiments due to its minimal parameter overhead and reasonable accuracy.
    
    \item \textbf{CHAOS Dataset (Figure \ref{fig8}c).} Rank 16 was selected based on its significant enhancement in Dice score (up to 0.8651 with 296 labels), coupled with maintaining a moderate model size. This balance made it a suitable candidate for subsequent fine-tuning and KD tasks.
    
    \item \textbf{PH2 Dataset (Figure \ref{fig8}d).} A LoRA rank of 8 provided high segmentation accuracy (0.9677 Dice score with 160 labels), making it optimal for downstream KD tasks given its efficiency and robust performance.
    
    \item \textbf{DRIVE Dataset (Figure \ref{fig8}e).} Rank 8 was employed for fine-tuning using only 20 labels, yielding a Dice score of 0.7598. This rank was deemed suitable for subsequent KD experiments, ensuring efficiency while maintaining a high level of segmentation performance.
\end{itemize}

\paragraph{\textbf{LoRA Rank Efficiency and Task-Specific SAM Fine-tuning.}}

Following the LoRA rank experiments, the selected ranks were used to fine-tune the SAM model with minimal labeled data for each task, thus serving as a task-specific SAM Teacher in the knowledge distillation framework (as shown in Table \ref{tab2}). For instance, in the DRIVE dataset, Rank 8 was chosen due to its efficiency in achieving strong segmentation with fewer parameters, providing an optimal balance for KD experiments.

The LoRA rank experiments allowed us to identify configurations that maximized segmentation performance while minimizing computational requirements. These fine-tuned models were subsequently employed as task-specific teachers within our knowledge distillation framework, ensuring that student models received guidance from highly optimized, task-specific teachers. This methodology facilitated an efficient and effective transfer of knowledge, ultimately enhancing the performance of the overall segmentation framework.
For more detailed experimental results of different rank Lora fine-tuning, see Table~\ref{taba6}- Table\ref{taba10}.

\begin{table*}[!ht]
\setlength{\abovecaptionskip}{0pt}
\setlength{\belowcaptionskip}{0pt}
\centering
% \small
% \normalsize
\large
\caption{Effect of LoRA Rank on segmentation performance for SAM \& LoRA Fine-Tuning on kidneyUS dataset with maximum labels.}
\scalebox{0.49}{
\begin{tabular}{cc
S[round-precision=4,round-mode=places]
S[round-precision=4,round-mode=places]
S[round-precision=4,round-mode=places]
S[round-precision=4,round-mode=places]
S[round-precision=4,round-mode=places]
S[round-precision=4,round-mode=places]
S[round-precision=4,round-mode=places]
S[round-precision=4,round-mode=places]
S[round-precision=4,round-mode=places]
S[round-precision=4,round-mode=places]}
\hline
\multirow{2}{*}{\textbf{Methods}} & \multirow{2}{*}{\textbf{Lora rank}} & \multicolumn{2}{c}{\textbf{Capsule}}   & \multicolumn{2}{c}{\textbf{Medulla}}   & \multicolumn{2}{c}{\textbf{Central Echo Complex}} & \multicolumn{2}{c}{\textbf{Cortex}}    & \multicolumn{2}{c}{\textbf{Mean}}      \\  
                                  &                                     & \textbf{Dice}     & \textbf{HD95 (mm)} & \textbf{Dice}     & \textbf{HD95 (mm)} & \textbf{Dice}           & \textbf{HD95 (mm)}      & \textbf{Dice}     & \textbf{HD95 (mm)} & \textbf{Dice}     & \textbf{HD95 (mm)} \\ \hline
SAM\&Lora Fine-Tuning             & 1                                   & 0.555704          & 6.003909           & 0.758542          & 6.510595           & 0.61568                 & 17.26695                & 0.521577          & 35.403537          & 0.612876          & 16.296248          \\
SAM\&Lora Fine-Tuning             & 2                                   & 0.522042          & 6.740435           & 0.770419          & 5.021455           & 0.621224                & 11.187499               & 0.530576          & 32.19494           & 0.611065          & 13.786082          \\
\rowcolor{lightblue}
SAM\&Lora Fine-Tuning             & \textbf{4}                          & 0.560059 & 5.674546  & 0.769286 & 6.374413  & 0.623433       & 13.984144      & 0.548723 & 27.131024 & 0.625375 & 13.291032 \\
SAM\&Lora Fine-Tuning             & 8                                   & 0.532384          & 5.273137           & 0.756023          & 4.910962           & 0.596544                & 17.647184               & 0.494088          & 44.054476          & 0.59476           & 17.97144           \\
SAM\&Lora Fine-Tuning             & 16                                  & 0.557161          & 5.192403           & 0.775705          & 4.224621           & 0.614365                & 15.849084               & 0.525001          & 41.880734          & 0.618058          & 16.786711          \\
SAM\&Lora Fine-Tuning             & 32                                  & 0.524804          & 5.262501           & 0.767873          & 5.559105           & 0.615706                & 16.768352               & 0.506354          & 24.287204          & 0.603684          & 12.969291          \\
SAM\&Lora Fine-Tuning             & 64                                  & 0.566021          & 5.77946            & 0.765535          & 4.9086             & 0.657547                & 7.226288                & 0.545545          & 22.530414          & 0.633662          & 10.11119           \\
SAM\&Lora Fine-Tuning             & 128                                 & 0.589576          & 4.699731           & 0.797141          & 3.985641           & 0.660266                & 11.552948               & 0.572901          & 26.205209          & 0.654971          & 11.610882          \\
SAM\&Lora Fine-Tuning             & 256                                 & 0.597963          & 4.719048           & 0.792125          & 4.250179           & 0.670837                & 13.202398               & 0.542627          & 23.744593          & 0.650888          & 11.479055          \\ \hline
\label{taba6}
\end{tabular}}
\vspace{-.1in}
\end{table*}

\begin{table*}[!ht]
\setlength{\abovecaptionskip}{0pt}
\setlength{\belowcaptionskip}{0pt}
\centering
% \small
% \normalsize
\large
\caption{Effect of LoRA Rank on segmentation performance for SAM \& LoRA Fine-Tuning on Autooral dataset with maximum labels.}
\scalebox{0.66}{
\begin{tabular}{cc
S[round-precision=4,round-mode=places]
S[round-precision=4,round-mode=places]
S[round-precision=4,round-mode=places]}
\hline
\textbf{Methods}       & \textbf{Lora rank} & \textbf{Dice}     & \textbf{HD95 (mm)} & \textbf{mIou}     \\ \hline
SAM\&Lora Fine-Tuning      & 1                  & 0.744602          & 6.996472           & 0.598233          \\
\rowcolor{lightblue}
SAM\&Lora Fine-Tuning                            & \textbf{2}                  & 0.812124 & 4.186127  & 0.687051 \\
SAM\&Lora Fine-Tuning                            & 4         & 0.799114          & 4.750000           & 0.668771          \\
SAM\&Lora Fine-Tuning                            & 8                  & 0.805574          & 4.579446           & 0.676524          \\
SAM\&Lora Fine-Tuning                            & 16                 & 0.801561          & 5.402750           & 0.672916          \\
SAM\&Lora Fine-Tuning                            & 32                 & 0.761061          & 5.927119           & 0.624327          \\
SAM\&Lora Fine-Tuning                            & 64                 & 0.787699          & 6.736173           & 0.652708          \\
SAM\&Lora Fine-Tuning                            & 128                & 0.777663          & 6.394663           & 0.645524          \\
SAM\&Lora Fine-Tuning                            & 256                & 0.787013          & 5.695578           & 0.652061          \\ \hline
\label{taba7}
\end{tabular}}
\vspace{-.1in}
\end{table*}

\begin{table*}[!ht]
\setlength{\abovecaptionskip}{0pt}
\setlength{\belowcaptionskip}{0pt}
\centering
% \small
% \normalsize
\large
\caption{Effect of LoRA Rank on segmentation performance for SAM \& LoRA Fine-Tuning on Chaos dataset with maximum labels.}
\scalebox{0.47}{
\begin{tabular}{cc
S[round-precision=4,round-mode=places]
S[round-precision=4,round-mode=places]
S[round-precision=4,round-mode=places]
S[round-precision=4,round-mode=places]
S[round-precision=4,round-mode=places]
S[round-precision=4,round-mode=places]
S[round-precision=4,round-mode=places]
S[round-precision=4,round-mode=places]
S[round-precision=4,round-mode=places]
S[round-precision=4,round-mode=places]}
\hline
\multirow{2}{*}{\textbf{Methods}} & \multirow{2}{*}{\textbf{Lora rank}} & \multicolumn{2}{c}{\textbf{Liver}}     & \multicolumn{2}{c}{\textbf{Right kidney}} & \multicolumn{2}{c}{\textbf{Left kidney}} & \multicolumn{2}{c}{\textbf{Spleen}}    & \multicolumn{2}{c}{\textbf{Mean}}      \\  
                                  &                                     & \textbf{Dice}     & \textbf{HD95 (mm)} & \textbf{Dice}       & \textbf{HD95 (mm)}  & \textbf{Dice}      & \textbf{HD95 (mm)}  & \textbf{Dice}     & \textbf{HD95 (mm)} & \textbf{Dice}     & \textbf{HD95 (mm)} \\ \hline
SAM\&Lora Fine-Tuning             & 1                                   & 0.886406          & 1.746433           & 0.856331            & 8.83603             & 0.825494           & 2.017055            & 0.82578           & 2.300313           & 0.848503          & 3.724958           \\
SAM\&Lora Fine-Tuning             & 2                                   & 0.892143          & 1.974668           & 0.871275            & 16.293329           & 0.861185           & 1.775955            & 0.845286          & 2.05504            & 0.867472          & 5.524748           \\
SAM\&Lora Fine-Tuning             & 4                                   & 0.893501          & 1.637578           & 0.846074            & 9.312596            & 0.853033           & 3.446246            & 0.851475          & 2.038706           & 0.861021          & 4.108782           \\
SAM\&Lora Fine-Tuning             & 8                                   & 0.895423          & 1.57072            & 0.852029            & 11.587411           & 0.843247           & 2.386545            & 0.835468          & 3.937531           & 0.856542          & 4.870551           \\
\rowcolor{lightblue} 
SAM\&Lora Fine-Tuning             & \textbf{16}                         & 0.890187 & 1.750513  & 0.855463   & 4.926637   & 0.875093  & 1.626032   & 0.839745 & 3.445731  & 0.865122 & 2.937228  \\
SAM\&Lora Fine-Tuning             & 32                                  & 0.888945          & 1.829311           & 0.872024            & 11.498926           & 0.878794           & 1.501271            & 0.840497          & 2.542181           & 0.870065          & 4.342922           \\
SAM\&Lora Fine-Tuning             & 64                                  & 0.896246          & 1.684418           & 0.871683            & 11.678663           & 0.884233           & 1.818182            & 0.875922          & 1.861098           & 0.882021          & 4.26059            \\
SAM\&Lora Fine-Tuning             & 128                                 & 0.909314          & 1.376933           & 0.884421            & 8.690108            & 0.884592           & 1.490641            & 0.866736          & 1.937751           & 0.886266          & 3.373858           \\
SAM\&Lora Fine-Tuning             & 256                                 & 0.912259          & 1.422753           & 0.897951            & 11.375587           & 0.899118           & 1.476604            & 0.874269          & 1.386894           & 0.895899          & 3.915459           \\ \hline
\label{taba8}
\end{tabular}}
\vspace{-.1in}
\end{table*}

\begin{table*}[!ht]
\setlength{\abovecaptionskip}{0pt}
\setlength{\belowcaptionskip}{0pt}
\centering
\large
\caption{Effect of LoRA Rank on segmentation performance for SAM \& LoRA Fine-Tuning on PH2 dataset with maximum labels. }
\scalebox{0.66}{
\begin{tabular}{cc
S[round-precision=4,round-mode=places]
S[round-precision=4,round-mode=places]
S[round-precision=4,round-mode=places]}
\hline
\textbf{Methods}        & \textbf{Lora rank} & \textbf{Dice}     & \textbf{HD95 (mm)} & \textbf{mIou}     \\ \hline
SAM\&Lora Fine-Tuning        & 1                  & 0.938082          & 1.901388           & 0.885213          \\
SAM\&Lora Fine-Tuning                            & 2                  & 0.940138          & 1.207107           & 0.888810          \\
SAM\&Lora Fine-Tuning                            & 4                  & 0.961623          & 0.750000           & 0.926219          \\
\rowcolor{lightblue}
SAM\&Lora Fine-Tuning                            & \textbf{8}         & 0.967723 & 0.500000  & 0.937507 \\
SAM\&Lora Fine-Tuning                            & 16                 & 0.962437          & 0.750000           & 0.927742          \\
SAM\&Lora Fine-Tuning                            & 32                 & 0.963592          & 0.603553           & 0.929901          \\
SAM\&Lora Fine-Tuning                            & 64                 & 0.964805          & 0.500000           & 0.932126          \\
SAM\&Lora Fine-Tuning                            & 128                & 0.965170          & 0.500000           & 0.932788          \\
SAM\&Lora Fine-Tuning                            & 256                & 0.968193          & 0.250000           & 0.938452          \\ \hline
\label{taba9}
\end{tabular}}
\vspace{-.1in}
\end{table*}

\begin{table*}[!ht]
\setlength{\abovecaptionskip}{0pt}
\setlength{\belowcaptionskip}{0pt}
\centering
% \small
% \normalsize
\large
\caption{Effect of LoRA Rank on segmentation performance for SAM \& LoRA Fine-Tuning on DRIVE dataset with maximum labels.}
\scalebox{0.66}{
\begin{tabular}{cc
S[round-precision=4,round-mode=places]
S[round-precision=4,round-mode=places]
S[round-precision=4,round-mode=places]}
\hline
\textbf{Methods}        & \textbf{Lora rank} & \textbf{Dice}     & \textbf{HD95 (mm)} & \textbf{mIou}     \\ \hline
SAM\&Lora Fine-Tuning       & 1                  & 0.733356          & 2.532248           & 0.578986          \\
SAM\&Lora Fine-Tuning                           & 2                  & 0.746165          & 2.224745           & 0.595150          \\
SAM\&Lora Fine-Tuning                           & 4                  & 0.739418          & 2.618034           & 0.586766          \\
\rowcolor{lightblue}
SAM\&Lora Fine-Tuning                            & \textbf{8}         & 0.759780 & 2.118034  & 0.612650 \\
SAM\&Lora Fine-Tuning                            & 16                 & 0.758867          & 2.118034           & 0.611464          \\
SAM\&Lora Fine-Tuning                           & 32                 & 0.755846          & 2.224745           & 0.607595          \\
SAM\&Lora Fine-Tuning                            & 64                 & 0.759125          & 2.118034           & 0.611783          \\
SAM\&Lora Fine-Tuning                           & 128                & 0.762373          & 2.118034           & 0.616017          \\
SAM\&Lora Fine-Tuning                           & 256                & 0.762429          & 2.118034           & 0.616124          \\ \hline
\label{taba10}
\end{tabular}}
\vspace{-.1in}
\end{table*}

\subsection{Effectiveness of Diffusion-Based Synthetic Transfer Data} 

\begin{table}[!ht]
\setlength{\abovecaptionskip}{0pt}
\setlength{\belowcaptionskip}{0pt}
% \small
\normalsize
\caption{ Evaluation of generated transfer datasets using diffusion Models on multiple medical imaging tasks.}
% \scalebox{0.98}{
\begin{tabular}{ccc}
\hline
\textbf{Dataset (Transfer)} & \textbf{PSNR (dB)} & \textbf{MSE} \\ \hline
KidneyUS Transfer           & 27.7475            & 109.4084     \\
Autooral Transfer           & 27.9109            & 105.2627     \\
Chaos Transfer              & 27.6902            & 111.4721     \\
PH2 Transfer                & 27.8803            & 105.9977     \\
DRIVE Transfer              & 27.9510             & 104.4108     \\ \hline
\label{tab3}
\end{tabular}
\vspace{-.1in}
\end{table}

To evaluate the quality of synthetic transfer datasets generated using diffusion models, we assessed the generated images compared to the original datasets using Peak Signal-to-Noise Ratio (PSNR) and Mean Squared Error (MSE) metrics. The evaluation, presented in Table \ref{tab3}, covered a range of medical imaging tasks, including KidneyUS, Autooral, CHAOS, PH2, and DRIVE. PSNR and MSE are standard metrics for assessing image quality, where higher PSNR values denote better image quality, and lower MSE values indicate closer similarity to the original images. Across all datasets, the diffusion model generated high-quality images, with PSNR values consistently around 27-28 dB and MSE values in the range of 104-111. This confirms that the generated images are visually similar to the original data, indicating that the synthetic datasets can serve as effective transfer sets for subsequent pre-training and fine-tuning tasks.

Notably, there was relatively little variation in the PSNR and MSE values across different datasets, suggesting that the diffusion model maintained a high and consistent quality of generated images regardless of the medical imaging task. For instance, the Autooral Transfer dataset achieved a PSNR of 27.91 dB and an MSE of 105.26, while the DRIVE Transfer dataset achieved similar results, with a PSNR of 27.95 dB and an MSE of 104.41. This consistency across diverse medical imaging modalities, such as ultrasound, retinal fundus, and dermoscopic images, demonstrates the generalizability of the diffusion model in generating reliable synthetic data for various medical segmentation tasks. The high PSNR and low MSE values validate the generated datasets' suitability for use in transfer learning, particularly in scenarios with limited labeled data. By utilizing these synthetic datasets for pre-training, the models benefited from enhanced performance through knowledge distillation and fine-tuning, ensuring that the synthetic data effectively captured the structural and visual features necessary for medical image segmentation tasks. Overall, the results affirm that the diffusion model can generate high-quality synthetic datasets suitable for addressing data scarcity in medical imaging tasks by augmenting original datasets with visually consistent synthetic images.

\section{Discussion and Conclusion}
\label{sec:conclusion}

The results of our study highlight the effectiveness of task-specific knowledge distillation (KD)~\cite{hinton2015distilling} and synthetic data generation for enhancing the performance of lightweight Vit models (such as ViT-Tiny) in medical image segmentation tasks. 
By utilizing a fine-tuned Vision Foundation Model (VFM) as a teacher, we successfully transferred domain-specific knowledge to a smaller Vision Transformer (ViT) model, leading to consistent performance gains across diverse datasets, especially in data-limited scenarios. This was particularly evident in challenging cases like retinal vessel and skin lesion segmentation, where the fine-tuned SAM model provided critical insights that enhanced the student's ability to capture fine-grained structures.

Our dual-level distillation strategy, aligning both encoder representations and decoder output logits, effectively facilitated the transfer of robust feature representations and semantic priors. This proved invaluable in overcoming the challenges associated with limited labeled data, where high-quality annotations are costly and time-consuming to obtain. Furthermore, Low-Rank Adaptation (LoRA) played a crucial role in efficiently fine-tuning the SAM model for specific tasks. The choice of LoRA rank balanced model performance with computational efficiency, with higher ranks being beneficial for complex datasets, while lower ranks sufficed for simpler tasks. This adaptability makes LoRA an attractive option for deploying medical AI systems in resource-constrained environments.

Synthetic data generation also proved beneficial, providing diverse pre-training data that closely resembled the original datasets. The use of Swin-transformer-based diffusion models for generating synthetic images enabled robust augmentation across various medical imaging modalities. Increasing the size of the transfer dataset generally improved model performance, though diminishing returns were noted beyond a certain threshold, emphasizing the need for a balanced approach to data generation.

In comparison to self-supervised pretraining methods like MoCo v3~\cite{chen2021empirical} and MAE~\cite{he2022masked}, task-specific KD demonstrated superior performance, especially when labeled data was scarce. Self-supervised methods often fall short of capturing the detailed structural information required for medical segmentation, which limits their utility in tasks where domain-specific knowledge is crucial. In contrast, task-specific KD leverages pre-trained models to infuse rich semantic information tailored to the task, making it highly effective for medical applications.

Our approach, combining pretraining on synthetic data with task-specific KD, also demonstrated efficiency in training. Despite requiring more computational resources than some self-supervised methods, the performance gains justified the added complexity. The training pipeline allowed for pre-training on synthetic data, followed by efficient fine-tuning, providing a scalable solution for real-world medical segmentation tasks, where timely and accurate decision-making is paramount.

However, several limitations warrant further research. The reliance on synthetic data may introduce biases that affect generalization to real clinical settings. Future studies should aim to validate these findings across broader, real-world datasets and explore alternative augmentation techniques. Additionally, the trade-offs between model size and segmentation accuracy should be further examined, potentially exploring other architectures or distillation methods that could offer similar or better performance without additional computational burden. Finally, applying our approach to more complex, multi-class segmentation tasks is an exciting direction for future work, as this could further showcase the benefits of transferring fine-grained anatomical knowledge.

In conclusion, our study underscores the potential of task-specific KD, combined with LoRA fine-tuning and synthetic data generation, to improve medical image segmentation, particularly in data-scarce environments significantly. This approach strikes an optimal balance between performance and computational efficiency, offering a viable pathway for scalable deployment in clinical practice, such as handling varying imaging modalities and adapting to new segmentation tasks. As the demand for efficient medical image analysis grows, the ability to adapt large models to specific tasks through distillation and fine-tuning will be increasingly crucial for advancing medical AI systems.

\vspace{40pt}
\section*{Data Availability}
The datasets utilized in this study are all publicly accessible. The KidneyUS~\cite{singla2023open}, Autooral~\cite{jiang2024high}, CHAOS~\cite{kavur2021chaos}, PH2~\cite{mendoncca2015ph2}, and DRIVE~\cite{staal2004ridge} datasets can be obtained from their respective sources, each providing detailed annotations suited to their specific medical imaging tasks. These datasets support reproducibility and further exploration of segmentation techniques in diverse medical imaging modalities.

\bibliography{sn-bibliography}

\end{document}